\documentclass[dvipsnames,format=sigconf,anonymous=false,review=false]{acmart}
\AtBeginDocument{%
  }

\setcopyright{acmlicensed}
\copyrightyear{2026}
\acmYear{2026}
\setcopyright{cc}
\setcctype{by-nc-nd}
\acmConference[GECCO '26]{Genetic and Evolutionary Computation Conference}{July 13--17, 2026}{San Jose, Costa Rica}
\acmBooktitle{Genetic and Evolutionary Computation Conference (GECCO '26), July 13--17, 2026, San Jose, Costa Rica}
\acmDOI{10.1145/3795095.3805067}
\acmISBN{979-8-4007-2487-9/2026/07}

\usepackage{microtype}
\usepackage{graphicx}
\usepackage{subcaption}
\usepackage{booktabs} 
\usepackage{hyperref}

\usepackage{algorithm}
\usepackage{algorithmic}
\usepackage{amsmath}
\usepackage{mathtools}
\usepackage{amsthm}
\usepackage[frozencache,cachedir=.]{minted}
\usepackage{adjustbox}
\usepackage{makecell}
\usepackage{libertine}

\newcommand{\model}[1]{%
\texttt{\hyphenchar\font=`\- #1}%
}

\newcommand{\KwInput}[1]{\STATE{{\bfseries Input:} #1}}
\newcommand{\KwOutput}[1]{\STATE{{\bfseries Output:} #1}}
\newcommand{\KwParam}[1]{\STATE{{\bfseries Param:} #1}}
\newcommand{\KwReturn}[1]{\STATE{{\bfseries Return} #1}}

\begin{document}

\title{LLM-ODE: Data-driven Discovery of Dynamical Systems with Large Language Models}

\author{Amirmohammad Ziaei Bideh}
\email{aziaeibideh@gradcenter.cuny.edu}
\affiliation{%
  \institution{The Graduate Center, CUNY}
  \city{New York}
  \state{New York}
  \country{USA}
}

\author{Jonathan Gryak}
\email{jonathan.gryak@qc.cuny.edu}
\orcid{0000-0002-5125-7741}
\affiliation{%
  \institution{Queens College and The Graduate Center, CUNY}
  \city{New York}
  \state{New York}
  \country{USA}
}

\renewcommand{\shortauthors}{Ziaei Bideh and Gryak}

\begin{abstract}
  Discovering the governing equations of dynamical systems is a central problem across many scientific disciplines. As experimental data become increasingly available, automated equation discovery methods offer a promising data-driven approach to accelerate scientific discovery. Among these methods, genetic programming (GP) has been widely adopted due to its flexibility and interpretability. However, GP-based approaches often suffer from inefficient exploration of the symbolic search space, leading to slow convergence and suboptimal solutions. To address these limitations, we propose LLM-ODE, a large language model-aided model discovery framework that guides symbolic evolution using patterns extracted from elite candidate equations. By leveraging the generative prior of large language models, LLM-ODE produces more informed search trajectories while preserving the exploratory strengths of evolutionary algorithms. Empirical results on 91 dynamical systems show that LLM-ODE variants consistently outperform classical GP methods in terms of search efficiency and Pareto-front quality. Overall, our results demonstrate that LLM-ODE improves both efficiency and accuracy over traditional GP-based discovery and offers greater scalability to higher-dimensional systems compared to linear and Transformer-only model discovery methods.
\end{abstract}




\maketitle

\section{Introduction}

Many natural and engineered phenomena are governed by systems of differential equations. Identifying the underlying governing equations is fundamental to understanding system dynamics and enables accurate prediction, analysis, and control across scientific domains such as physics, biology, chemistry, and engineering. Traditionally, the discovery of such equations has relied heavily on domain expertise, physical intuition, and manual derivation. While effective in well-understood settings, this process is time-consuming, difficult to scale, and increasingly inadequate for modern data-rich scientific problems.

\begin{figure}[t]
    \centering
    \includegraphics[width=1\linewidth]{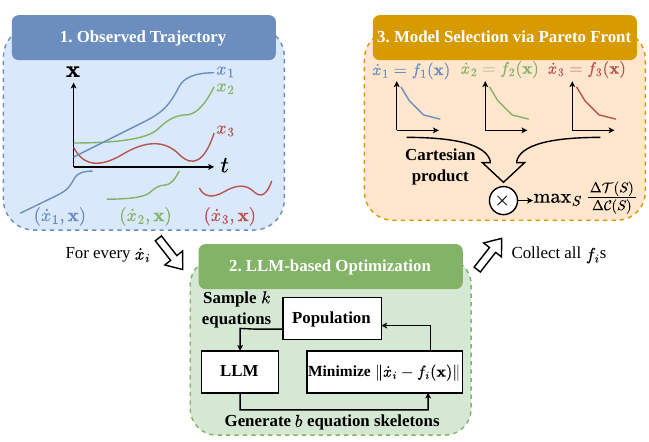}
    \caption{Schematic overview of LLM-ODE. (1) Given trajectory data from an unknown dynamical system, the observations are decomposed into state variables. (2) A large language model acts as an evolutionary operator, guiding the evolution of symbolic equation populations toward higher-fitness candidates. (3) The final system of equations is selected from the Cartesian product of equation-level Pareto fronts.}
    \label{fig:llmode}
\end{figure}

Motivated by the growing availability of experimental and simulation data, data-driven model discovery methods have emerged as a promising alternative for uncovering interpretable governing equations directly from observations. Among these approaches, genetic programming (GP) \cite{koza1994genetic} has been one of the most widely adopted tools for symbolic equation discovery due to its flexibility and expressiveness. However, GP-based methods typically rely on stochastic mutation and crossover operators, which often lead to inefficient exploration of the vast combinatorial search space of symbolic expressions \cite{kronberger2024inefficiency}. As a result, GP can suffer from slow convergence, limited scalability, and difficulty in discovering high-quality equations for complex or high-dimensional dynamical systems.

Recent advances in large language models (LLMs) suggest a new opportunity to address these challenges. Beyond their success in natural language processing \cite{turc2019} and code generation \cite{li2022competition}, LLMs have demonstrated strong capabilities in scientific reasoning and discovery \cite{wang2023scientific}, optimization \cite{yang2023large}, and black-box recombination \cite{meyerson2024language, lange2024large}. These models implicitly capture rich structural and syntactic patterns from large corpora, making them well suited to propose meaningful symbolic expressions and transformations. This raises a natural question: can LLMs be used to guide symbolic evolution toward more informative and efficient search trajectories?

In this work, we answer this question by introducing \textbf{LLM-ODE}, an LLM-guided genetic programming framework for discovering the governing equations of dynamical systems. As illustrated in Figure~\ref{fig:llmode}, LLM-ODE replaces conventional stochastic evolutionary operators, such as random mutation and crossover, with LLM-generated proposals that exploit patterns observed in high-performing candidate equations. Given observed trajectory data from an unknown dynamical system, LLM-ODE decomposes the system into its state variables and evolves symbolic expressions for each governing equation using LLM-guided operators. Candidate equations are evaluated using multi-objective criteria, and final systems are selected from the Cartesian product of equation-level Pareto fronts.

While prior work has explored LLM-based symbolic regression for discovering single equations, to the best of our knowledge this is the first method that leverages LLMs within an evolutionary framework to discover \textit{systems of coupled differential equations} describing dynamical systems. By combining the generative prior of LLMs with the diversity of evolutionary search, LLM-ODE aims to achieve more efficient and scalable model discovery.

Our main contributions are summarized as follows:
\begin{itemize}
    \item We introduce a new LLM-aided framework that leverages the strength of LLMs and evolutionary search to discover the governing equations of dynamical systems.
    \item We conduct a comprehensive empirical evaluation on a benchmark suite of 91 dynamical systems, spanning one to four state variables and including chaotic dynamics.
    \item We show that LLM-ODE, across multiple LLM variants, consistently outperforms traditional GP in terms of search efficiency and system Pareto-front quality.
\end{itemize}

\section{Related Work}

Symbolic regression (SR) methods aim to infer interpretable equations directly from data. More broadly, model discovery methods seek to uncover analytical expressions that govern nonlinear dynamical systems. In contrast with other ML algorithms, SR methods enjoy several advantages \cite{dong2025recent}: 1) they do not need a predefined model structure, 2) they automatically generate concise, human-readable equations modeling the relationship between data, and 3) they often exhibit stronger extrapolation performance, a property more pronounced in scarce-data settings \cite{wilstrup2021symbolic}.

Model discovery approaches can be categorized into several groups. \textbf{Linear Models (LM)}, pioneered by SINDy \cite{brunton2016discovering} and its extensions, e.g., \citet{messenger2021weak}, work with the assumption that the governing equations are a linear combination of predefined basic functions with sparse scalar coefficients. Such methods impose a strong prior on the equation space -- linearity in library terms -- which can be useful in discovering simple models with few involving variables. These methods are limited by the need for specifying data-specific candidate library terms, which hinders the scalability of the scientific discovery task.

\textbf{GP} methods relax the linearity assumption and aim to search among the expression trees that best fit the data. These methods, such as \citet{cranmer2023interpretable} and \citet{burlacu2020operon}, iteratively evolve a pool of expression trees through evolutionary variations between equations while maintaining high-performing equations. It has been shown that GP methods remain among the most accurate approaches for model discovery \cite{bideh2025mdbenchbenchmarkingdatadrivenmethods, la2021contemporary, aldeia2025call}. One of the limitations of GP-based methods for SR is their inefficiency in exploring the search space, mainly due to the repeated evaluation of semantically equivalent equations during the search process \cite{kronberger2024inefficiency}.

\textbf{Large-scale pretraining (LSPT)} algorithms are Transformer models \cite{vaswani2017attention} that are pre-trained on a large synthetic SR dataset. The pre-trained model auto-regressively generates symbolic equations directly from new data points \cite{biggio2021neural, kamienny2022end, d2024odeformer}. While some methods optimize the constants in the equations with an external component \cite{biggio2021neural}, in other methods, the Transformer directly produces complete mathematical equations \cite{d2024odeformer}. More recent work adds search components such as Monte Carlo Tree Search to LSPT methods to improve the transformer equation generation process \cite{shojaee2023transformer}.

\textbf{LLM-aided} methods have flourished from a recently developed paradigm for scientific discovery \cite{romera2024mathematical}, where LLMs' optimization \cite{yang2023large} and in-context learning abilities \cite{meyerson2024language} are leveraged as intelligent variational operators \cite{brown2020language}. LLM-SR \cite{shojaee2025llm} aims to uncover the governing equations given a natural language description of the scientific problem. \citet{merler2024context} omits the equation descriptions and includes data points directly in the LLM input. \cite{du2024large} introduces an LLM-aided GP framework for model discovery of nonlinear dynamics, limited to dynamics solely with one state variable. More recent work allows more LLM flexibility by promoting the LLM as an autonomous AI scientist that can perform explanatory data analysis for scientific discovery \cite{xia2025sr}.

For a more comprehensive analysis of model discovery methods, see \citet{aldeia2025call, bideh2025mdbenchbenchmarkingdatadrivenmethods, dong2025recent}.

\section{Methodology}

\subsection{Problem Statement}
We consider the problem of discovering governing equations of an unknown dynamical system from observed trajectory data. Let $\mathbf{X} \in \mathbb{R}^{N \times D}$ denote a time series of $N$ samples of a $D$-dimensional state $\mathbf{x}(t) = (x_1(t), \dots, x_D(t))$. The system is assumed to evolve according to a set of autonomous ordinary differential equations (ODEs)
\[
\dot{\mathbf{x}}_i = f_i(\mathbf{x}), \quad i = 1,\dots,D,
\]
where each $f_i : \mathbb{R}^D \rightarrow \mathbb{R}$ is an unknown symbolic expression.

Our goal is to recover a system
\[
S := \{f_1, \dots, f_D\}
\]
	
that accurately reproduces the observed dynamics while remaining compact and interpretable. To formalize this trade-off, we define the complexity of an expression $f$ as the number of nodes in its expression tree, denoted $\mathcal{C}(f)$. The complexity of a system is then
\[
\mathcal{C}(S) = \sum_{i=1}^D \mathcal{C}(f_i).
\]
We seek models that jointly minimize prediction error and symbolic complexity.

\subsection{Overview of LLM-ODE}

LLM-ODE is a hybrid symbolic regression framework that combines GP with LLMs to improve search efficiency in the space of symbolic expressions. Classical GP-based symbolic regression methods often suffer from inefficient exploration of the exponentially large space of expression trees, leading to slow convergence and premature stagnation \cite{kronberger2024inefficiency}. LLM-ODE addresses this limitation by using an LLM to propose structurally informed mutations and recombinations, leveraging patterns observed in high-performing equations \cite{brown2020language, mirchandani2023large}.

The algorithm follows an island-based evolutionary strategy \cite{cranmer2023interpretable}, where multiple populations evolve in parallel. Each island maintains a population of candidate symbolic expressions and evolves independently for several iterations, promoting exploration of diverse regions of the search space.

Rather than searching directly over full systems of equations, LLM-ODE decomposes the problem across state variables. For each state variable $x_i$, the algorithm independently learns a symbolic approximation of its time derivative $f_i(\mathbf{x})$. This decomposition reduces the combinatorial complexity of the search and allows targeted optimization of each equation.

Algorithms~\ref{alg:llmode1}, \ref{alg:llmode2}, and \ref{alg:llmode3} summarize the pseudocode of \textsc{LLM-ODE}. Each population is initialized with two randomly generated simple equations of complexity three, and is subsequently evolved in an iterative manner.

At each iteration, $k$ individuals are sampled from the current population according to a softmax distribution proportional to $\exp(s)$, where $s$ denotes the fitness score computed with Equation \ref{eq:score}. The selected individuals are used as in-context examples to prompt the LLM, which proposes $b$ candidate expressions intended to improve upon the sampled equations.

For each candidate, the symbolic structure is fixed while its numerical constants and coefficients are optimized on the training data using an external numerical optimizer. The resulting expressions are then evaluated, and the optimized candidates are inserted back into the population for subsequent evolution.

Each island evolves independently and may therefore over-specialize in a restricted region of the equation search space. To alleviate this effect and promote broader exploration, we introduce a periodic refinement step. After every \( n_{\text{refine}} \) evolutionary rounds, \( n_{\text{mix}} \) equations from each island are randomly exchanged with another island, after which the lower-performing half of each population is pruned. This mechanism encourages information sharing across islands while maintaining selection pressure, thereby improving diversity and preventing premature convergence.

After completing \( n_{\text{iter}} \) iterations, the algorithm returns the Pareto front of individuals with respect to symbolic complexity and prediction error. In this setting, the Pareto front consists of all expressions \( f_i \) such that there exists no \( f_j \) satisfying
\[
\mathcal{C}(f_j) \leq \mathcal{C}(f_i)
\quad \text{and} \quad
\mathcal{S}(f_j) > \mathcal{S}(f_i),
\]
where \( \mathcal{C}(\cdot) \) denotes symbolic complexity and \( \mathcal{S}(\cdot) \) denotes the fitness score. The fitness score is defined as negative mean squared error 

\begin{equation}
\label{eq:score}
\mathcal{S}(f_i) = -\text{MSE} \left ( \dot{\mathbf{x}}, f_i \left (\mathbf{x} \right ) \right ).
\end{equation}

To construct a final system of ordinary differential equations, we further aggregate candidate equations into systems and compute a second Pareto front over systems, trading off total complexity and trajectory-level accuracy. System-level fitness is defined as the negative mean squared error (MSE) between predicted and observed trajectories on a validation set,
\[
\mathcal{T}(S) = -\mathrm{MSE}\big(S(\mathbf{x}_0), \mathbf{x}\big),
\]
where \( S(\mathbf{x}_0) \) denotes the trajectory obtained by numerically integrating system \( S \) from the observed initial condition \( \mathbf{x}_0 \).

To balance model complexity and predictive accuracy, we select the final discovered system from the system-level Pareto front
\( \mathbf{S} = \{ S_1, \dots, S_M \} \) by identifying the point of maximum marginal gain in trajectory fitness per unit increase in complexity. Specifically, we solve
\begin{equation}
\arg\max_{i \in \{2,\dots,M\}}
\frac{\mathcal{T}(S_i) - \mathcal{T}(S_{i-1})}
{\mathcal{C}(S_i) - \mathcal{C}(S_{i-1})}
\quad \text{s.t.} \quad
\mathcal{T}(S_i) \geq h\,\mathcal{T}_{\max},
\label{eq:min}
\end{equation}
where \( \mathcal{T}_{\max} = \max_i \mathcal{T}(S_i) \). The threshold \( h = 0.1 \) prevents premature selection of overly simple but inaccurate systems by restricting the search to sufficiently well-performing candidates.

\begin{algorithm}[h]
\caption{LLM-ODE}
\label{alg:llmode1}
\begin{algorithmic}[1]
  \KwInput{ODE trajectories $\mathcal{X}$ = $\{\mathcal{X_\text{train}}, \mathcal{X_\text{val}}\}$}
  \KwOutput{optimal system of equations $S^*$}
  \KwParam{minimum accuracy threshold $h$}
  
    \STATE{$n \leftarrow$ number of state variables}
    \STATE{$E \leftarrow \{\}$ \texttt{// Pareto fronts}}
    
    \FOR{$i\leftarrow 1$ to $n$}
        \STATE{$y_\text{train}, y_\text{val} \leftarrow $ approximate $dx_i/dt$} for $\mathcal{X_\text{train}}, \mathcal{X_\text{val}}$, respectively
        \STATE{$E' \leftarrow$ LLMSymReg$(\mathcal{X}_\text{train}, \mathcal{X}_\text{val}, y_\text{train}, y_\text{val})$}
        \STATE{$E \leftarrow E \cup \{E'\}$}
    \ENDFOR

    \STATE{\textit{systemPool} $\leftarrow \text{cartesianProduct}(E)$ \texttt{// pool of system of equations from all the combinations of discovered equations}} 

    \STATE{ \textit{systemPF} $\leftarrow$ System Pareto front of \textit{systemPool} } 
    
    \STATE{$S^* \leftarrow \arg \max_{S \in \text{\textit{systemPF}}} \frac{\Delta\mathcal{T}(S)}{\Delta \mathcal{C}(S)} \;\text{s.t.}\; \mathcal{T}(S) \ge h\mathcal{T}_\text{max} $}
    
    \KwReturn{$S^*$}
\end{algorithmic}
\end{algorithm}

\begin{algorithm}[h]
\begin{algorithmic}[1]
  \KwInput{$\mathcal{X}_\text{train}, \mathcal{X}_\text{val}, y_\text{train}, y_\text{val}$}
  \KwOutput{equations along the Pareto front $E$}
  \KwParam{number of populations or islands $n_{p}$, number of iterations $n_{\text{iter}}$, refinement frequency $n_\text{refine}$, number of equations to mix $n_\text{mix}$}
  
    \STATE{\texttt{// initialize seed equations}}
    \FOR{$i\leftarrow 1$ \textbf{to} $n_p$}
        \STATE{$P_i \leftarrow 2 \ \text{random, simple equations}$} 
        \FOR{$p$ \textbf{in} $P_i$}
            \STATE{Optimize constants in $p$ to minimize $\text{MSE}(y_\text{train}, p(\mathcal{X_\text{train}}))$}
        \ENDFOR
    \ENDFOR
    \FOR{$n \leftarrow 1$ \textbf{to} $n_\mathrm{iter}$}
        \FOR{$i \leftarrow 1$ \textbf{to} $n_p$}
            \STATE{$P_i \leftarrow \mathrm{evolve}(P_i, \mathcal{X}_{\text{train}}, y^{(i)}_\text{train})$}
        \ENDFOR
        \IF{$n \mod n_\text{refine} = 0 $}
            \STATE{Prune all islands $\cup_i P_i$}
            \STATE{Sample $n_\text{mix}$ equations from every island and add to another random island}
        \ENDIF
    \ENDFOR
    \STATE{$P \leftarrow \cup_i P_i$}
    \STATE{$E \leftarrow$ Pareto front of $p \in P$ on  $\mathcal{X}_\text{val}, y_\text{val}$}
    
    \KwReturn{E}
\end{algorithmic}
\caption{LLMSymReg (LLM-aided symbolic regression)}
\label{alg:llmode2}
\end{algorithm}

\begin{algorithm}[h]
\begin{algorithmic}[1]
  \KwInput{an island consisting of single equations $P$, observed trajectory $\mathcal{X}_\text{train}$, target values $y$}
  \KwOutput{evolved island $P$}
  \KwParam{LLM $\pi$, number of in-context examples $k$, number of samples per prompt $b$}
  
    \STATE{\textit{score} $\leftarrow p \in P : -\text{MSE}(y, p(\mathcal{X_\text{train}}))$}
    \STATE{$w \leftarrow \frac{e^{\textit{score}}}{\sum e^\text{score}}$}
    \STATE{\normalfont $Q \leftarrow $ sample $k$ equations from island $P$ weighted by $w$}
    \STATE{$Q' \leftarrow$ sample $b$ new equations from the LLM $\pi$ given in-context examples $Q$}
    \STATE{optimize constants in $p \in Q'$ to minimize MSE$(y, p(\mathcal{X_\text{train}}))$}
    \STATE{$P \leftarrow P \cup Q'$}
    \KwReturn{P}
\end{algorithmic}
\caption{evolve (LLM-aided Evolution)}
\label{alg:llmode3}
\end{algorithm}

\section{Experiments}

\subsection{LLM-ODE Setup}
\begin{listing}[h]
\begin{minted}[fontsize=\small, breaklines]{json}
// LLM input
[
  {
    "role": "system",
    "content": "You are a scientist whose task is to perform Symbolic Regression. You should search the function space to find the best simple function that fits the data. You are given 8 examples of proposed equations sorted from worst to best. Your goal is to suggest 3 improved equations of varying complexity. Replace all numerical constants with "C" -- they will be optimized with an external optimizer. Write one equation per line from simplest to most complex with no extra explanation. Available operators: +, -, *, **, /, sin, log, exp, abs. Independent variables: x_0, x_1, x_2."
  },
  {
    // Proposed equations from prior iterations
    "role": "user",
    "content": 
    "x_1**C
    C*x_0 + x_1
    C*sin(x_2)
    C*log(exp(x_2) + Abs(x_1))
    C*x_0 + C*sin(x_2)
    C*x_0*x_1
    (C + Abs(x_2))**C*(C*x_0**C + C*exp(x_1))
    C*log(x_0 + Abs(x_1)) + C*sin(x_2)"
  }
]

// LLM output
[
  {
    "role": "assistant",
    "content": 
    "C*x_0 + C*log(Abs(x_1) + 1)
    C*(x_0 + sin(x_2)) * exp(x_1)
    C*(x_0**2 + x_1**2) / (1 + abs(x_2))"
  }
]]
\end{minted}
\caption{Example input prompt with $k=8$ in-context examples and $b=3$ generated equations.}
\label{lis:sample}
\end{listing}

As noted in prior work \cite{shojaee2025llmsrbench}, a key challenge in evaluating LLM-aided methods for scientific discovery is the \emph{memorization problem}, whereby well-known scientific equations may have been encountered during the LLM’s pretraining phase. To mitigate this issue, we remove any information about the semantic meaning of state variables and the underlying physical characteristics of the dynamical system from the input prompts. This design choice encourages the LLM to rely on structural reasoning and pattern matching \cite{mirchandani2023large}, rather than direct recall of memorized equations.

To account for variability across language models, we evaluate LLM-ODE using three different LLM backends: \model{Qwen3-30B-A3B-Instruct-2507} (Qwen)  \cite{qwen3technicalreport}, \model{Ministral-3-14B-Instruct-2512} (Mistral) \cite{liu2026ministral3}, and \model{Olmo-3.1-32B-Instruct} (Olmo) \cite{olmo2025olmo}. All models are served using vLLM, which provides high-throughput inference via efficient key-value cache management with PagedAttention \cite{kwon2023efficient}.

The LLM-ODE framework employs four independent evolutionary islands, each initialized with two seed expressions. Each island is evolved for 200 iterations, with the number of in-context examples set to $k=8$ and the number of samples per prompt set to $b=3$. The refinement procedure is applied every $n_{\text{refine}} = 5$ iterations.

The input to the framework consists of a single observed trajectory along with estimated time derivatives. Throughout this work, we compute derivatives using a fourth-order finite-difference scheme \cite{findiff}. These derivatives are used to optimize constant placeholders in candidate equations via the BFGS algorithm \cite{broyden1970convergence}. Discovered systems of ordinary differential equations are then simulated using SciPy’s ODE solver \cite{2020SciPy-NMeth}, which employ explicit Runge-Kutta integration. The set of admissible mathematical operators is restricted to $\{+, -, /, \text{\textasciicircum}, \sin, \log, \exp, \mathrm{abs}\}$ and is explicitly specified in the input prompt. A representative prompt and corresponding LLM-generated output are provided in Listing~\ref{lis:sample}.

To promote reproducibility and facilitate future research, we release the full source code for dataset generation, LLM-ODE, and all baseline methods in a publicly available GitHub repository\footnote{\url{https://github.com/gryaklab/llm-ode}}.

\begin{table*}
\caption{Number of discovered systems by method based on test trajectory accuracy. A successful discovery corresponds to a system with test trajectory NMSE below $\lambda$. The dataset is stratified with respect to the number of state variables $D$, followed by the number of relevant systems. In almost all cases, LLM-ODE methods achieve a higher or equal discovery rate compared to PySR, suggesting the superiority of LLM-based search to uninformed search.}
\label{tab:discovery}
\centering
\setlength{\tabcolsep}{3pt}
\begin{adjustbox}{max width=\textwidth}
\begin{tabular}{lcccccc|cccccc|cccccc|cccccc}
\toprule
Method
& \multicolumn{6}{c}{D = 1 ($n=23$)}
& \multicolumn{6}{c}{D = 2 ($n=28$)}
& \multicolumn{6}{c}{D = 3 ($n=22$)}
& \multicolumn{6}{c}{D = 4 ($n=18$)} \\
\cmidrule(lr){2-7}
\cmidrule(lr){8-13}
\cmidrule(lr){14-19}
\cmidrule(lr){20-25}
   \multicolumn{1}{r}{$\lambda$} & $10^{-1}$ & $10^{-2}$ & $10^{-3}$ & $10^{-4}$ & $10^{-5}$ & $10^{-6}$
& $10^{-1}$ & $10^{-2}$ & $10^{-3}$ & $10^{-4}$ & $10^{-5}$ & $10^{-6}$
& $10^{-1}$ & $10^{-2}$ & $10^{-3}$ & $10^{-4}$ & $10^{-5}$ & $10^{-6}$
& $10^{-1}$ & $10^{-2}$ & $10^{-3}$ & $10^{-4}$ & $10^{-5}$ & $10^{-6}$ \\
\midrule

LLM-ODE (Mistral)
& 20 & 19 & 16 & 15 & \textbf{13} & \textbf{8}
& 15 & 10 & 7 & 6 & 3 & 1
& 3 & 3 & 3 & \textbf{2} & \textbf{1} & 0
& 4 & \textbf{4} & \textbf{3} & \textbf{3} & \textbf{3} & \textbf{3} \\

LLM-ODE (Olmo)
& \textbf{21} & 19 & \textbf{18} & \textbf{16} & 11 & \textbf{8}
& \textbf{18} & 12 & 10 & \textbf{9} & 5 & 1
& 4 & 2 & 0 & 0 & 0 & 0
& \textbf{6} & \textbf{4} & 1 & 1 & 1 & 0 \\

LLM-ODE (Qwen)
& \textbf{21} & \textbf{20} & 17 & \textbf{16} & \textbf{13} & \textbf{8}
& 17 & 11 & 9 & 4 & 4 & \textbf{3}
& 8 & 6 & 4 & \textbf{2} & 0 & 0
& 4 & 2 & 2 & 1 & 0 & 0 \\

PySR
& 20 & \textbf{20} & 16 & 13 & 8 & 6
& 16 & 9 & 6 & 6 & 2 & 1
& 4 & 4 & 3 & 1 & \textbf{1} & 0
& 1 & 1 & 0 & 0 & 0 & 0 \\ \hline

ODEFormer
& 18 & 16 & 10 & 4 & 2 & 0
& 12 & 5 & 1 & 0 & 0 & 0
& 2 & 1 & 0 & 0 & 0 & 0
& 1 & 0 & 0 & 0 & 0 & 0 \\

SINDy
& 20 & 15 & 12 & 6 & 5 & 3
& \textbf{18} & \textbf{16} & \textbf{12} & \textbf{9} & \textbf{6} & \textbf{3}
& \textbf{10} & \textbf{8} & \textbf{5} & 1 & 0 & 0
& 1 & 0 & 0 & 0 & 0 & 0 \\

\bottomrule
\end{tabular}
\end{adjustbox}
\end{table*}

\subsection{Baseline Methods}

In order to assess the significance of the improvement in model discovery using the proposed framework, we evaluate three different methods from distinct algorithmic classes to establish a baseline. \textbf{PySR} \cite{cranmer2023interpretable} is an efficient implementation of a GP algorithm. This method serves as a baseline for a regular GP algorithm with no heuristic used in guiding the search space exploration. \textbf{SINDy} \cite{brunton2016discovering} and \textbf{ODEFormer} \cite{d2024odeformer} are included as representative approaches from LM and LSPT methods, respectively. 

To make a fair comparison, we have endeavored to make the training process for all the baseline models similar to LLM-ODE. PySR performs SR on each state variable separately and computes a set of complexity-error Pareto fronts for each equation in the system. We follow the same procedure as LLM-ODE to construct the proposed system of equations and balance the tradeoff.

We train PySR for 200 iterations with 4 islands, 3 cycles per iterations, 20 individuals, and the same set of allowed operators as in LLM-ODE. The SINDy model is trained using the STLSQ algorithm with all the combinations in the hyperparameter search space, which is shown in Table \ref{tab:sindy-hyper} in Appendix. For ODEFormer, we follow the authors' guidelines and set beam size to 50 and use a temperature of 0.1. We consider the 50 generated systems resulting from beam search as the Pareto front. All baseline models use Equation \ref{eq:min} to pick the final discovered system of equations along the system Pareto front.

\subsection{Dataset}
\label{sec:dataset}

We evaluate LLM-ODE alongside several baseline methods on a comprehensive suite of 91 dynamical systems, all simulated from real-world models. Our experiments begin with the ODEBench dataset \cite{d2024odeformer} ($n=63$), which provides a diverse set of systems for benchmarking. Upon inspection, we observed that the ODEBench dataset is heavily skewed toward systems with fewer state variables, limiting its coverage of high-dimensional and chaotic dynamics. To address this gap, we extend the dataset with additional systems exhibiting more complex behaviors, resulting in a total of 23 systems with $D=1$, 28 with $D=2$, 22 with $D=3$, and 18 with $D=4$. Detailed specifications of the equations and their parameters are provided in Appendix \ref{sec:app-data} and in the code supplementary material.

For each system, we generate two distinct trajectories corresponding to different initial conditions. One trajectory is used for training, while the other serves as the test set. All trajectories are integrated over the time interval (measured in seconds) $t = 0$ to $t = 10$ using a uniform sampling interval of 0.1 seconds. In our experiments, we use the trajectory segment from $t = 0$ to $t = 5$ for parameter optimization. The full training trajectory is then employed for model selection to ensure robust evaluation across different dynamical regimes.

\subsection{Results}
\begin{figure*}[h]
    \centering
    \includegraphics[width=1\linewidth]{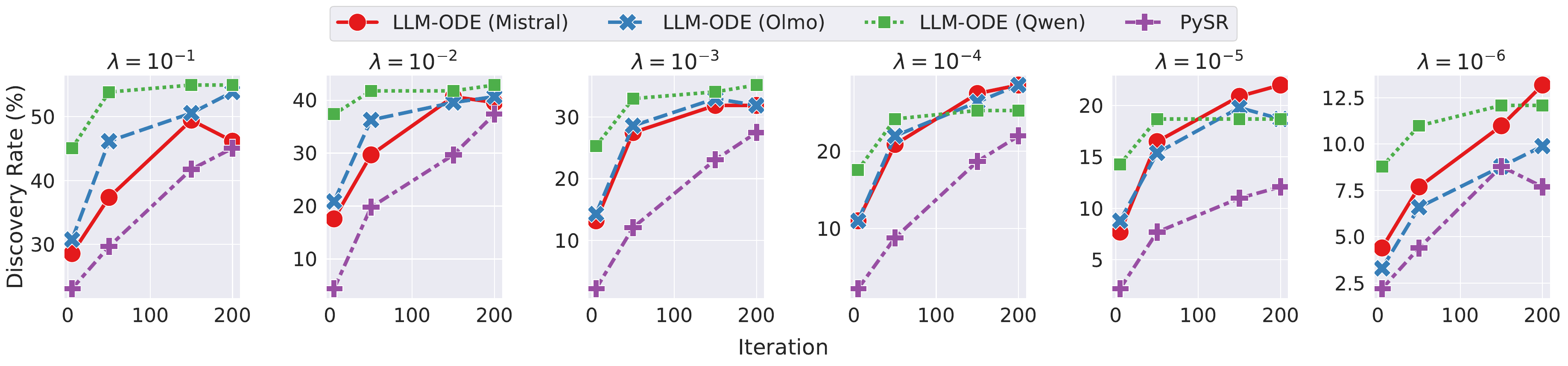}
    \caption{System discovery rate as a function of search iterations across various NMSE thresholds $\lambda$. Overall, LLM-ODE methods achieve higher discovery rates across all iteration limits. LLM-ODE methods enjoy a faster convergence rate than PySR.}
    \label{fig:convergence}
\end{figure*}

\begin{figure}[h]
    \centering
    \includegraphics[width=1\linewidth]{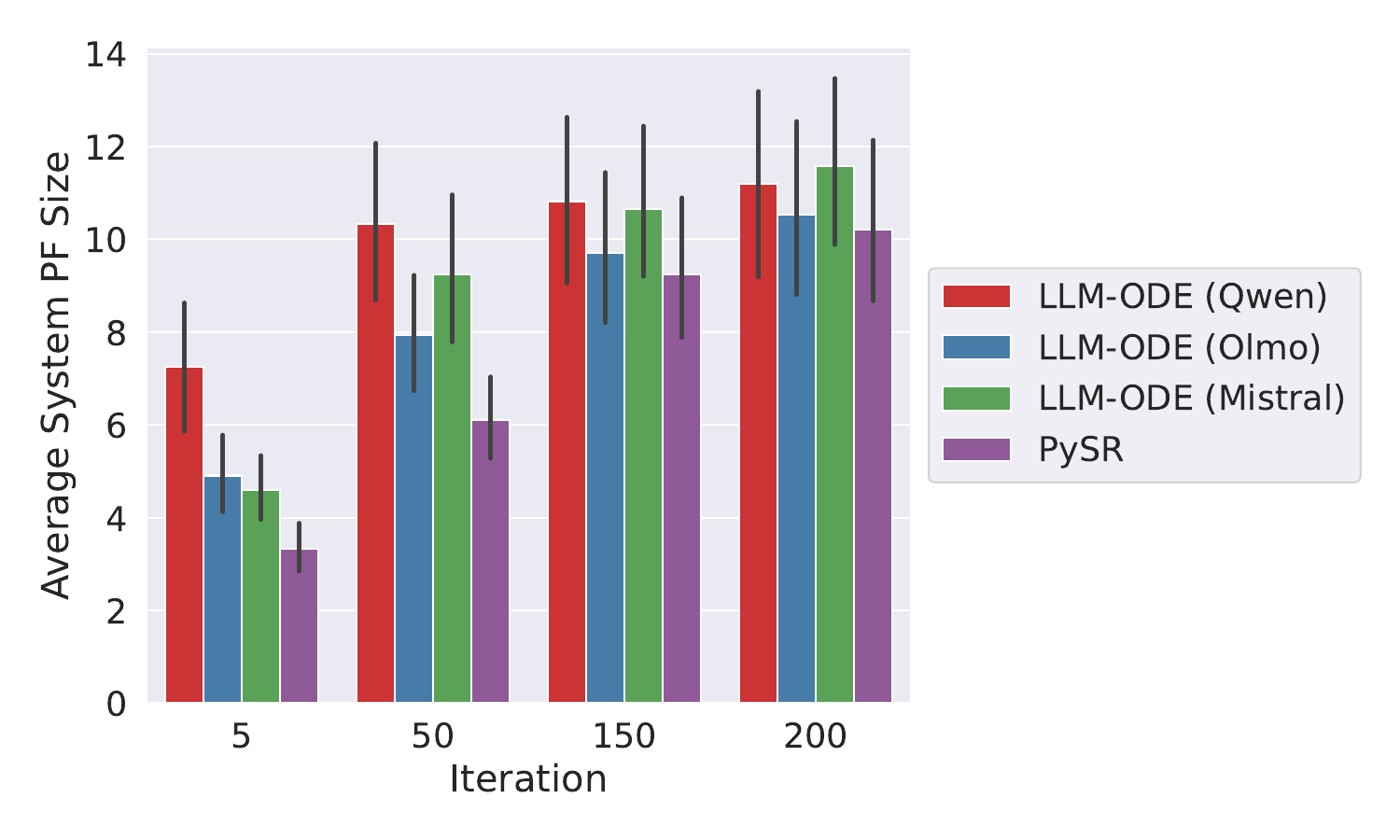}
    \caption{The number of points along the system Pareto front of GP-based methods during training. The error bars represent 95\% confidence intervals across systems. LLM-ODE methods are more efficient at obtaining a diverse Pareto-front as compared to stochastic GP.}
    \label{fig:average-pf-number}
\end{figure}

\subsubsection{System Discovery}

Table \ref{tab:discovery} reports the number of systems successfully discovered by each method as a function of the normalized mean squared error (NMSE) threshold $\lambda$, stratified by system dimensionality $D$. A successful discovery is defined as achieving a test trajectory NMSE below $\lambda$. We use NMSE rather than MSE to normalize for differences in the scale of state variables across systems, enabling fairer comparisons between systems of varying magnitudes.

Across all dimensions, performance degrades monotonically as the error tolerance becomes stricter (smaller $\lambda$), which is expected and validates the consistency of the evaluation. Lower-dimensional systems ($D=1$ and $D=2$) are substantially easier to recover than higher-dimensional ones ($D=3$ and $D=4$), indicating a sharp increase in discovery difficulty with state dimension.


LLM-ODE methods consistently outperform ODEFormer across all system classes and error thresholds. This observation suggests that purely Transformer-based approaches may be limited in their ability to directly infer precise governing equations from raw observational data. In contrast, hybrid frameworks that integrate learned priors with an explicit symbolic search component appear to yield more accurate and reliable system discovery.

For low-dimensional dynamical systems with a single state variable ($D=1$), LLM-ODE achieves consistently higher discovery rates than PySR and SINDy across a wide range of error thresholds. In intermediate dimensions ($D=2,3$), SINDy attains competitive, and in some regimes higher, discovery rates relative to other methods. However, this trend does not persist in higher dimensions: for $D=4$, SINDy’s performance collapses, while LLM-ODE variants continue to successfully recover multiple systems. The degradation of SINDy in higher dimensions is likely due to the exponential growth of candidate library terms and the resulting difficulty of sparse regression. In contrast, LLM-ODE methods remain effective and discover substantially more systems than stochastic genetic programming approaches, highlighting their superior scalability in complex search spaces.

These results demonstrate that LLM-guided search substantially improves symbolic system discovery, particularly under strict accuracy constraints and increasing system complexity. The consistent advantage of LLM-ODE over PySR suggests that leveraging language-model priors enables more effective exploration of the hypothesis space than uninformed symbolic search. 

\subsubsection{Convergence Rate}

Figure \ref{fig:convergence} illustrates the discovery rate of GP-based methods as a function of search iterations across various successful discovery thresholds. Across all tested precision thresholds, LLM-ODE methods consistently outperform PySR in both total discovery rate and convergence speed. Notably, LLM-ODE variants achieve a significantly higher discovery rate within the first 50 iterations than PySR achieves after 200 iterations in several scenarios (e.g., at $\lambda = 10^{-3}$ and $\lambda = 10^{-5}$). This suggests that leveraging the prior knowledge and reasoning capabilities of LLMs as evolutionary operators provides a more intelligent and efficient search compared to the uninformed search mechanisms used in classic GP methods.

As the discovery threshold $\lambda$ becomes stricter (moving from $10^{-1}$ to $10^{-6}$), the absolute discovery rate for all methods naturally decreases. However, the performance gap between LLM-ODE and PySR remains robust. At high tolerance regimes ($\lambda \in \{10^{-1}, 10^{-2}, 10^{-3}\}$), LLM-ODE (Qwen) maintains a dominant lead, reaching near-plateau discovery rates much earlier than other models. At low tolerance thresholds ($\lambda \in \{10^{-5}, 10^{-6}\}$), LLM-ODE (Mistral) demonstrates remarkable resilience, eventually overtaking other LLM variants and maintaining a clear margin over PySR, which struggles to exceed a 10\% discovery rate at $\lambda = 10^{-6}$.

\subsubsection{Pareto Front Analysis}

From a multi-objective optimization perspective, the number and distribution of points along the Pareto front are widely used as proxies for search quality. A larger, well-distributed Pareto front implies better coverage of the objective space and greater diversity in the underlying population. In genetic programming and evolutionary computation, maintaining a diverse set of non-dominated individuals is known to improve robustness, reduce premature convergence, and increase the probability of discovering globally competitive solutions \cite{audet2021performance}.

Figure \ref{fig:average-pf-number} compares the evolution of Pareto fronts obtained by LLM-ODE variants and PySR over training iterations for all the systems in the dataset. Each Pareto front captures the trade-off between model complexity and validation error, and thus provides a direct view into the structure and quality of the explored equation search space.

In the initial training steps, the population evolved by LLMs show a substantially richer Pareto front compared to PySR. However, this gap shrinks as the evolution continues, to the point that no significant difference is observed. This observation suggests that intelligent search operations require fewer search iterations to achieve a rich, diverse Pareto front, whereas uninformed GP methods are slower and less efficient in achieving the same level of Pareto front richness. 

The observed Pareto front richness of LLM-ODE therefore suggests a more effective exploration of the equation search space. By leveraging LLM-generated priors to guide symbolic search, LLM-ODE appears to explore multiple semantically distinct regions of the space simultaneously, rather than relying solely on stochastic variation. This results not only in better best-case solutions, but also in a broader set of viable models that reflect different structural hypotheses.

\subsubsection{Limitations}

The proposed framework relies on an LLM to generate candidate equations during the GP steps. The latency of LLM inference introduces a computational bottleneck, resulting in slower execution compared to non-LLM GP methods. Moreover, running LLMs typically requires specialized hardware that may not be accessible to all users. Quantized and low-rank approximation LLM models offer faster and more lightweight alternatives, broadening the accessibility of the framework at the cost of modest performance degradation.

\begin{figure}
    \centering
    \includegraphics[width=0.9\linewidth]{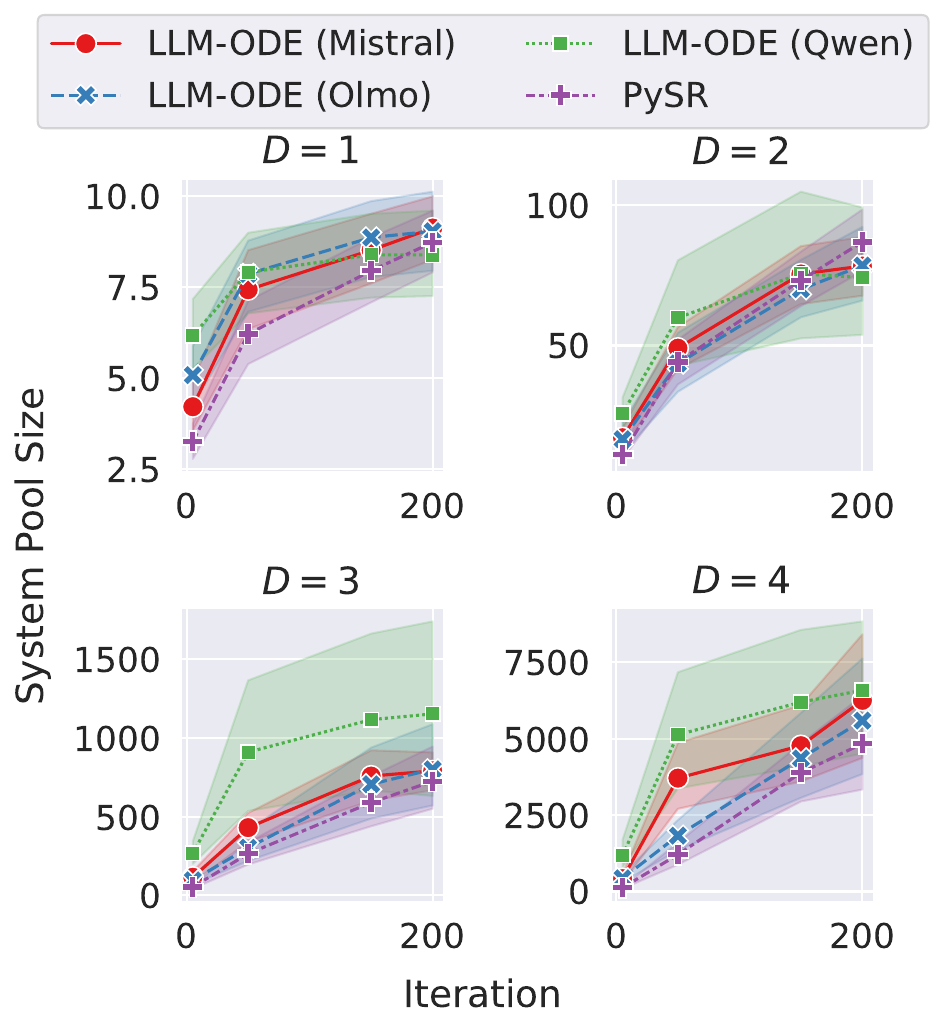}
    \caption{System pool sizes from which the system Pareto-front is computed.}
    \label{fig:sys-pool}
\end{figure}

Although the proposed method demonstrates better scalability to high-dimensional systems compared to classical GP, both approaches struggle in very high-dimensional settings. Figure \ref{fig:sys-pool} illustrates the system pool sizes, defined as the product of the equation Pareto-front sizes across individual equations. At the final iteration, the system Pareto-front is computed from this pool. At a system size of 4, the pool contains $O(10^4)$ candidate systems, which remains computationally manageable. Nevertheless, this highlights a general limitation of GP-based methods: their scalability degrades significantly as system dimensionality increases. As computing the system Pareto-front is a computationally expensive operation it is deferred to the final iteration, after the GP loop has concluded.

\section{Conclusion}

This work presents LLM-ODE, a novel framework that integrates large language models with genetic programming for discovering governing equations of dynamical systems from observational data. By replacing traditional stochastic evolutionary operators with LLM-guided suggestions, our approach achieves more intelligent exploration of the equation search space. Empirical evaluation on 91 dynamical systems demonstrates that LLM-ODE consistently outperforms classical genetic programming methods and Transformer-only approaches across multiple dimensions of system complexity.

Our results reveal several key insights. First, LLM-guided search substantially improves both discovery rate and convergence speed, particularly for low-dimensional systems where LLM-ODE variants discover accurate equations within 50 iterations, a milestone that traditional GP methods fail to reach even after 200 iterations.

Second, the LLM-aided GP variants result in more diverse system Pareto front in fewer steps compared to uninformed GP, suggesting that leveraging language model priors enables more effective hypothesis space exploration than uninformed symbolic search.

Third, LLM-ODE demonstrates better scalability to systems of higher dimensionality compared to not only the stochastic GP and Transformer-based methods, but also linear model-based approaches like SINDy, which collapse in four-dimensional settings while LLM-ODE continues to successfully recover multiple systems. The remaining performance gap with ground truth solutions in higher dimensions indicates that scalability remains an open challenge; however, the gains over other model discovery methods across all precision thresholds support LLM-based approaches as a promising direction for robust, scalable model discovery.

In this work, to mitigate potential information leakage and prevent LLMs from simply exploiting memorized equations, we deliberately exclude all system-level metadata from the input prompts, including the system description, state variable names, and physical units. Access to such information could allow the LLMs to leverage embedded scientific knowledge and accelerate convergence toward accurate equations in fewer evolution steps. As a direction for future work, the LLM-ODE framework can be applied to genuinely novel scientific systems, where system information is made available to the backend LLM, enabling the discovery of previously unknown models.

\bibliographystyle{ACM-Reference-Format}
\bibliography{refs}


\begin{thebibliography}{38}


\ifx \showCODEN    \undefined \def \showCODEN     #1{\unskip}     \fi
\ifx \showISBNx    \undefined \def \showISBNx     #1{\unskip}     \fi
\ifx \showISBNxiii \undefined \def \showISBNxiii  #1{\unskip}     \fi
\ifx \showISSN     \undefined \def \showISSN      #1{\unskip}     \fi
\ifx \showLCCN     \undefined \def \showLCCN      #1{\unskip}     \fi
\ifx \shownote     \undefined \def \shownote      #1{#1}          \fi
\ifx \showarticletitle \undefined \def \showarticletitle #1{#1}   \fi
\ifx \showURL      \undefined \def \showURL       {\relax}        \fi
\providecommand\bibfield[2]{#2}
\providecommand\bibinfo[2]{#2}
\providecommand\natexlab[1]{#1}
\providecommand\showeprint[2][]{arXiv:#2}

\bibitem[Aldeia et~al\mbox{.}(2025)]%
        {aldeia2025call}
\bibfield{author}{\bibinfo{person}{Guilherme S~Imai Aldeia},
  \bibinfo{person}{Hengzhe Zhang}, \bibinfo{person}{Geoffrey Bomarito},
  \bibinfo{person}{Miles Cranmer}, \bibinfo{person}{Alcides Fonseca},
  \bibinfo{person}{Bogdan Burlacu}, \bibinfo{person}{William~G La~Cava}, {and}
  \bibinfo{person}{Fabr{\'\i}cio~Olivetti de Fran{\c{c}}a}.}
  \bibinfo{year}{2025}\natexlab{}.
\newblock \showarticletitle{Call for Action: towards the next generation of
  symbolic regression benchmark}.
\newblock \bibinfo{journal}{\emph{arXiv preprint arXiv:2505.03977}}
  (\bibinfo{year}{2025}).
\newblock


\bibitem[Audet et~al\mbox{.}(2021)]%
        {audet2021performance}
\bibfield{author}{\bibinfo{person}{Charles Audet}, \bibinfo{person}{Jean
  Bigeon}, \bibinfo{person}{Dominique Cartier}, \bibinfo{person}{S{\'e}bastien
  Le~Digabel}, {and} \bibinfo{person}{Ludovic Salomon}.}
  \bibinfo{year}{2021}\natexlab{}.
\newblock \showarticletitle{Performance indicators in multiobjective
  optimization}.
\newblock \bibinfo{journal}{\emph{European journal of operational research}}
  \bibinfo{volume}{292}, \bibinfo{number}{2} (\bibinfo{year}{2021}),
  \bibinfo{pages}{397--422}.
\newblock


\bibitem[Baer(2018)]%
        {findiff}
\bibfield{author}{\bibinfo{person}{M. Baer}.} \bibinfo{year}{2018}\natexlab{}.
\newblock \bibinfo{title}{{findiff} Software Package}.
\newblock
\urldef\tempurl%
\url{https://github.com/maroba/findiff}
\showURL{%
\tempurl}
\newblock
\shownote{\url{https://github.com/maroba/findiff}}.


\bibitem[Bideh et~al\mbox{.}(2025)]%
        {bideh2025mdbenchbenchmarkingdatadrivenmethods}
\bibfield{author}{\bibinfo{person}{Amirmohammad~Ziaei Bideh},
  \bibinfo{person}{Aleksandra Georgievska}, {and} \bibinfo{person}{Jonathan
  Gryak}.} \bibinfo{year}{2025}\natexlab{}.
\newblock \bibinfo{title}{MDBench: Benchmarking Data-Driven Methods for Model
  Discovery}.
\newblock
\showeprint[arxiv]{2509.20529}~[cs.LG]
\urldef\tempurl%
\url{https://arxiv.org/abs/2509.20529}
\showURL{%
\tempurl}


\bibitem[Biggio et~al\mbox{.}(2021)]%
        {biggio2021neural}
\bibfield{author}{\bibinfo{person}{Luca Biggio}, \bibinfo{person}{Tommaso
  Bendinelli}, \bibinfo{person}{Alexander Neitz}, \bibinfo{person}{Aurelien
  Lucchi}, {and} \bibinfo{person}{Giambattista Parascandolo}.}
  \bibinfo{year}{2021}\natexlab{}.
\newblock \showarticletitle{Neural symbolic regression that scales}. In
  \bibinfo{booktitle}{\emph{International Conference on Machine Learning}}.
  Pmlr, \bibinfo{pages}{936--945}.
\newblock


\bibitem[Brown et~al\mbox{.}(2020)]%
        {brown2020language}
\bibfield{author}{\bibinfo{person}{Tom Brown}, \bibinfo{person}{Benjamin Mann},
  \bibinfo{person}{Nick Ryder}, \bibinfo{person}{Melanie Subbiah},
  \bibinfo{person}{Jared~D Kaplan}, \bibinfo{person}{Prafulla Dhariwal},
  \bibinfo{person}{Arvind Neelakantan}, \bibinfo{person}{Pranav Shyam},
  \bibinfo{person}{Girish Sastry}, \bibinfo{person}{Amanda Askell},
  {et~al\mbox{.}}} \bibinfo{year}{2020}\natexlab{}.
\newblock \showarticletitle{Language models are few-shot learners}.
\newblock \bibinfo{journal}{\emph{Advances in neural information processing
  systems}}  \bibinfo{volume}{33} (\bibinfo{year}{2020}),
  \bibinfo{pages}{1877--1901}.
\newblock


\bibitem[Broyden(1970)]%
        {broyden1970convergence}
\bibfield{author}{\bibinfo{person}{Charles~George Broyden}.}
  \bibinfo{year}{1970}\natexlab{}.
\newblock \showarticletitle{The convergence of a class of double-rank
  minimization algorithms 1. general considerations}.
\newblock \bibinfo{journal}{\emph{IMA Journal of Applied Mathematics}}
  \bibinfo{volume}{6}, \bibinfo{number}{1} (\bibinfo{year}{1970}),
  \bibinfo{pages}{76--90}.
\newblock


\bibitem[Brunton et~al\mbox{.}(2016)]%
        {brunton2016discovering}
\bibfield{author}{\bibinfo{person}{Steven~L Brunton}, \bibinfo{person}{Joshua~L
  Proctor}, {and} \bibinfo{person}{J~Nathan Kutz}.}
  \bibinfo{year}{2016}\natexlab{}.
\newblock \showarticletitle{Discovering governing equations from data by sparse
  identification of nonlinear dynamical systems}.
\newblock \bibinfo{journal}{\emph{Proceedings of the national academy of
  sciences}} \bibinfo{volume}{113}, \bibinfo{number}{15}
  (\bibinfo{year}{2016}), \bibinfo{pages}{3932--3937}.
\newblock


\bibitem[Burlacu et~al\mbox{.}(2020)]%
        {burlacu2020operon}
\bibfield{author}{\bibinfo{person}{Bogdan Burlacu}, \bibinfo{person}{Gabriel
  Kronberger}, {and} \bibinfo{person}{Michael Kommenda}.}
  \bibinfo{year}{2020}\natexlab{}.
\newblock \showarticletitle{Operon C++ an efficient genetic programming
  framework for symbolic regression}. In \bibinfo{booktitle}{\emph{Proceedings
  of the 2020 Genetic and Evolutionary Computation Conference Companion}}.
  \bibinfo{pages}{1562--1570}.
\newblock


\bibitem[Cranmer(2023)]%
        {cranmer2023interpretable}
\bibfield{author}{\bibinfo{person}{Miles Cranmer}.}
  \bibinfo{year}{2023}\natexlab{}.
\newblock \showarticletitle{Interpretable machine learning for science with
  PySR and SymbolicRegression. jl}.
\newblock \bibinfo{journal}{\emph{arXiv preprint arXiv:2305.01582}}
  (\bibinfo{year}{2023}).
\newblock


\bibitem[d'Ascoli et~al\mbox{.}(2024)]%
        {d2024odeformer}
\bibfield{author}{\bibinfo{person}{St{\'e}phane d'Ascoli},
  \bibinfo{person}{S{\"o}ren Becker}, \bibinfo{person}{Philippe Schwaller},
  \bibinfo{person}{Alexander Mathis}, {and} \bibinfo{person}{Niki Kilbertus}.}
  \bibinfo{year}{2024}\natexlab{}.
\newblock \showarticletitle{{ODEF}ormer: Symbolic Regression of Dynamical
  Systems with Transformers}. In \bibinfo{booktitle}{\emph{The Twelfth
  International Conference on Learning Representations}}.
\newblock
\urldef\tempurl%
\url{https://openreview.net/forum?id=TzoHLiGVMo}
\showURL{%
\tempurl}


\bibitem[Dong and Zhong(2025)]%
        {dong2025recent}
\bibfield{author}{\bibinfo{person}{Junlan Dong} {and} \bibinfo{person}{Jinghui
  Zhong}.} \bibinfo{year}{2025}\natexlab{}.
\newblock \showarticletitle{Recent Advances in Symbolic Regression}.
\newblock \bibinfo{journal}{\emph{Comput. Surveys}} \bibinfo{volume}{57},
  \bibinfo{number}{11} (\bibinfo{year}{2025}), \bibinfo{pages}{1--37}.
\newblock


\bibitem[Du et~al\mbox{.}(2024)]%
        {du2024large}
\bibfield{author}{\bibinfo{person}{Mengge Du}, \bibinfo{person}{Yuntian Chen},
  \bibinfo{person}{Zhongzheng Wang}, \bibinfo{person}{Longfeng Nie}, {and}
  \bibinfo{person}{Dongxiao Zhang}.} \bibinfo{year}{2024}\natexlab{}.
\newblock \showarticletitle{Large language models for automatic equation
  discovery of nonlinear dynamics}.
\newblock \bibinfo{journal}{\emph{Physics of Fluids}} \bibinfo{volume}{36},
  \bibinfo{number}{9} (\bibinfo{year}{2024}).
\newblock


\bibitem[Kamienny et~al\mbox{.}(2022)]%
        {kamienny2022end}
\bibfield{author}{\bibinfo{person}{Pierre-Alexandre Kamienny},
  \bibinfo{person}{St{\'e}phane d'Ascoli}, \bibinfo{person}{Guillaume Lample},
  {and} \bibinfo{person}{Fran{\c{c}}ois Charton}.}
  \bibinfo{year}{2022}\natexlab{}.
\newblock \showarticletitle{End-to-end symbolic regression with transformers}.
\newblock \bibinfo{journal}{\emph{Advances in Neural Information Processing
  Systems}}  \bibinfo{volume}{35} (\bibinfo{year}{2022}),
  \bibinfo{pages}{10269--10281}.
\newblock


\bibitem[Koza(1994)]%
        {koza1994genetic}
\bibfield{author}{\bibinfo{person}{John~R Koza}.}
  \bibinfo{year}{1994}\natexlab{}.
\newblock \showarticletitle{Genetic programming as a means for programming
  computers by natural selection}.
\newblock \bibinfo{journal}{\emph{Statistics and computing}}
  \bibinfo{volume}{4} (\bibinfo{year}{1994}), \bibinfo{pages}{87--112}.
\newblock


\bibitem[Kronberger et~al\mbox{.}(2024)]%
        {kronberger2024inefficiency}
\bibfield{author}{\bibinfo{person}{Gabriel Kronberger},
  \bibinfo{person}{Fabricio Olivetti~de Franca}, \bibinfo{person}{Harry
  Desmond}, \bibinfo{person}{Deaglan~J Bartlett}, {and} \bibinfo{person}{Lukas
  Kammerer}.} \bibinfo{year}{2024}\natexlab{}.
\newblock \showarticletitle{The inefficiency of genetic programming for
  symbolic regression}. In \bibinfo{booktitle}{\emph{International Conference
  on Parallel Problem Solving from Nature}}. Springer,
  \bibinfo{pages}{273--289}.
\newblock


\bibitem[Kwon et~al\mbox{.}(2023)]%
        {kwon2023efficient}
\bibfield{author}{\bibinfo{person}{Woosuk Kwon}, \bibinfo{person}{Zhuohan Li},
  \bibinfo{person}{Siyuan Zhuang}, \bibinfo{person}{Ying Sheng},
  \bibinfo{person}{Lianmin Zheng}, \bibinfo{person}{Cody~Hao Yu},
  \bibinfo{person}{Joseph Gonzalez}, \bibinfo{person}{Hao Zhang}, {and}
  \bibinfo{person}{Ion Stoica}.} \bibinfo{year}{2023}\natexlab{}.
\newblock \showarticletitle{Efficient memory management for large language
  model serving with pagedattention}. In \bibinfo{booktitle}{\emph{Proceedings
  of the 29th symposium on operating systems principles}}.
  \bibinfo{pages}{611--626}.
\newblock


\bibitem[La~Cava et~al\mbox{.}(2021)]%
        {la2021contemporary}
\bibfield{author}{\bibinfo{person}{William La~Cava}, \bibinfo{person}{Bogdan
  Burlacu}, \bibinfo{person}{Marco Virgolin}, \bibinfo{person}{Michael
  Kommenda}, \bibinfo{person}{Patryk Orzechowski},
  \bibinfo{person}{Fabr{\'\i}cio~Olivetti de Fran{\c{c}}a},
  \bibinfo{person}{Ying Jin}, {and} \bibinfo{person}{Jason~H Moore}.}
  \bibinfo{year}{2021}\natexlab{}.
\newblock \showarticletitle{Contemporary symbolic regression methods and their
  relative performance}.
\newblock \bibinfo{journal}{\emph{Advances in neural information processing
  systems}} \bibinfo{volume}{2021}, \bibinfo{number}{DB1}
  (\bibinfo{year}{2021}), \bibinfo{pages}{1}.
\newblock


\bibitem[Lange et~al\mbox{.}(2024)]%
        {lange2024large}
\bibfield{author}{\bibinfo{person}{Robert Lange}, \bibinfo{person}{Yingtao
  Tian}, {and} \bibinfo{person}{Yujin Tang}.} \bibinfo{year}{2024}\natexlab{}.
\newblock \showarticletitle{Large language models as evolution strategies}. In
  \bibinfo{booktitle}{\emph{Proceedings of the Genetic and Evolutionary
  Computation Conference Companion}}. \bibinfo{pages}{579--582}.
\newblock


\bibitem[Li et~al\mbox{.}(2022)]%
        {li2022competition}
\bibfield{author}{\bibinfo{person}{Yujia Li}, \bibinfo{person}{David Choi},
  \bibinfo{person}{Junyoung Chung}, \bibinfo{person}{Nate Kushman},
  \bibinfo{person}{Julian Schrittwieser}, \bibinfo{person}{R{\'e}mi Leblond},
  \bibinfo{person}{Tom Eccles}, \bibinfo{person}{James Keeling},
  \bibinfo{person}{Felix Gimeno}, \bibinfo{person}{Agustin Dal~Lago},
  {et~al\mbox{.}}} \bibinfo{year}{2022}\natexlab{}.
\newblock \showarticletitle{Competition-level code generation with alphacode}.
\newblock \bibinfo{journal}{\emph{Science}} \bibinfo{volume}{378},
  \bibinfo{number}{6624} (\bibinfo{year}{2022}), \bibinfo{pages}{1092--1097}.
\newblock


\bibitem[Liu et~al\mbox{.}(2026)]%
        {liu2026ministral3}
\bibfield{author}{\bibinfo{person}{Alexander~H. Liu}, \bibinfo{person}{Kartik
  Khandelwal}, \bibinfo{person}{Sandeep Subramanian}, \bibinfo{person}{Victor
  Jouault}, \bibinfo{person}{Abhinav Rastogi}, \bibinfo{person}{Adrien Sadé},
  \bibinfo{person}{Alan Jeffares}, \bibinfo{person}{Albert Jiang},
  \bibinfo{person}{Alexandre Cahill}, \bibinfo{person}{Alexandre Gavaudan},
  \bibinfo{person}{Alexandre Sablayrolles}, \bibinfo{person}{Amélie Héliou},
  \bibinfo{person}{Amos You}, \bibinfo{person}{Andy Ehrenberg},
  \bibinfo{person}{Andy Lo}, \bibinfo{person}{Anton Eliseev},
  \bibinfo{person}{Antonia Calvi}, \bibinfo{person}{Avinash Sooriyarachchi},
  \bibinfo{person}{Baptiste Bout}, \bibinfo{person}{Baptiste Rozière},
  \bibinfo{person}{Baudouin~De Monicault}, \bibinfo{person}{Clémence
  Lanfranchi}, \bibinfo{person}{Corentin Barreau}, \bibinfo{person}{Cyprien
  Courtot}, \bibinfo{person}{Daniele Grattarola}, \bibinfo{person}{Darius
  Dabert}, \bibinfo{person}{Diego de~las Casas}, \bibinfo{person}{Elliot
  Chane-Sane}, \bibinfo{person}{Faruk Ahmed}, \bibinfo{person}{Gabrielle
  Berrada}, \bibinfo{person}{Gaëtan Ecrepont}, \bibinfo{person}{Gauthier
  Guinet}, \bibinfo{person}{Georgii Novikov}, \bibinfo{person}{Guillaume
  Kunsch}, \bibinfo{person}{Guillaume Lample}, \bibinfo{person}{Guillaume
  Martin}, \bibinfo{person}{Gunshi Gupta}, \bibinfo{person}{Jan Ludziejewski},
  \bibinfo{person}{Jason Rute}, \bibinfo{person}{Joachim Studnia},
  \bibinfo{person}{Jonas Amar}, \bibinfo{person}{Joséphine Delas},
  \bibinfo{person}{Josselin~Somerville Roberts}, \bibinfo{person}{Karmesh
  Yadav}, \bibinfo{person}{Khyathi Chandu}, \bibinfo{person}{Kush Jain},
  \bibinfo{person}{Laurence Aitchison}, \bibinfo{person}{Laurent Fainsin},
  \bibinfo{person}{Léonard Blier}, \bibinfo{person}{Lingxiao Zhao},
  \bibinfo{person}{Louis Martin}, \bibinfo{person}{Lucile Saulnier},
  \bibinfo{person}{Luyu Gao}, \bibinfo{person}{Maarten Buyl},
  \bibinfo{person}{Margaret Jennings}, \bibinfo{person}{Marie Pellat},
  \bibinfo{person}{Mark Prins}, \bibinfo{person}{Mathieu Poirée},
  \bibinfo{person}{Mathilde Guillaumin}, \bibinfo{person}{Matthieu Dinot},
  \bibinfo{person}{Matthieu Futeral}, \bibinfo{person}{Maxime Darrin},
  \bibinfo{person}{Maximilian Augustin}, \bibinfo{person}{Mia Chiquier},
  \bibinfo{person}{Michel Schimpf}, \bibinfo{person}{Nathan Grinsztajn},
  \bibinfo{person}{Neha Gupta}, \bibinfo{person}{Nikhil Raghuraman},
  \bibinfo{person}{Olivier Bousquet}, \bibinfo{person}{Olivier Duchenne},
  \bibinfo{person}{Patricia Wang}, \bibinfo{person}{Patrick von Platen},
  \bibinfo{person}{Paul Jacob}, \bibinfo{person}{Paul Wambergue},
  \bibinfo{person}{Paula Kurylowicz}, \bibinfo{person}{Pavankumar~Reddy
  Muddireddy}, \bibinfo{person}{Philomène Chagniot}, \bibinfo{person}{Pierre
  Stock}, \bibinfo{person}{Pravesh Agrawal}, \bibinfo{person}{Quentin Torroba},
  \bibinfo{person}{Romain Sauvestre}, \bibinfo{person}{Roman Soletskyi},
  \bibinfo{person}{Rupert Menneer}, \bibinfo{person}{Sagar Vaze},
  \bibinfo{person}{Samuel Barry}, \bibinfo{person}{Sanchit Gandhi},
  \bibinfo{person}{Siddhant Waghjale}, \bibinfo{person}{Siddharth Gandhi},
  \bibinfo{person}{Soham Ghosh}, \bibinfo{person}{Srijan Mishra},
  \bibinfo{person}{Sumukh Aithal}, \bibinfo{person}{Szymon Antoniak},
  \bibinfo{person}{Teven~Le Scao}, \bibinfo{person}{Théo Cachet},
  \bibinfo{person}{Theo~Simon Sorg}, \bibinfo{person}{Thibaut Lavril},
  \bibinfo{person}{Thiziri~Nait Saada}, \bibinfo{person}{Thomas Chabal},
  \bibinfo{person}{Thomas Foubert}, \bibinfo{person}{Thomas Robert},
  \bibinfo{person}{Thomas Wang}, \bibinfo{person}{Tim Lawson},
  \bibinfo{person}{Tom Bewley}, \bibinfo{person}{Tom Bewley},
  \bibinfo{person}{Tom Edwards}, \bibinfo{person}{Umar Jamil},
  \bibinfo{person}{Umberto Tomasini}, \bibinfo{person}{Valeriia Nemychnikova},
  \bibinfo{person}{Van Phung}, \bibinfo{person}{Vincent Maladière},
  \bibinfo{person}{Virgile Richard}, \bibinfo{person}{Wassim Bouaziz},
  \bibinfo{person}{Wen-Ding Li}, \bibinfo{person}{William Marshall},
  \bibinfo{person}{Xinghui Li}, \bibinfo{person}{Xinyu Yang},
  \bibinfo{person}{Yassine~El Ouahidi}, \bibinfo{person}{Yihan Wang},
  \bibinfo{person}{Yunhao Tang}, {and} \bibinfo{person}{Zaccharie Ramzi}.}
  \bibinfo{year}{2026}\natexlab{}.
\newblock \bibinfo{title}{Ministral 3}.
\newblock
\showeprint[arxiv]{2601.08584}~[cs.CL]
\urldef\tempurl%
\url{https://arxiv.org/abs/2601.08584}
\showURL{%
\tempurl}


\bibitem[Merler et~al\mbox{.}(2024)]%
        {merler2024context}
\bibfield{author}{\bibinfo{person}{Matteo Merler}, \bibinfo{person}{Katsiaryna
  Haitsiukevich}, \bibinfo{person}{Nicola Dainese}, {and}
  \bibinfo{person}{Pekka Marttinen}.} \bibinfo{year}{2024}\natexlab{}.
\newblock \showarticletitle{In-Context Symbolic Regression: Leveraging Large
  Language Models for Function Discovery}. In
  \bibinfo{booktitle}{\emph{Proceedings of the 62nd Annual Meeting of the
  Association for Computational Linguistics (Volume 4: Student Research
  Workshop)}}, \bibfield{editor}{\bibinfo{person}{Xiyan Fu} {and}
  \bibinfo{person}{Eve Fleisig}} (Eds.). \bibinfo{publisher}{Association for
  Computational Linguistics}, \bibinfo{address}{Bangkok, Thailand},
  \bibinfo{pages}{427--444}.
\newblock
\showISBNx{979-8-89176-097-4}
\href{https://doi.org/10.18653/v1/2024.acl-srw.49}{doi:\nolinkurl{10.18653/v1/2024.acl-srw.49}}


\bibitem[Messenger and Bortz(2021)]%
        {messenger2021weak}
\bibfield{author}{\bibinfo{person}{Daniel~A Messenger} {and}
  \bibinfo{person}{David~M Bortz}.} \bibinfo{year}{2021}\natexlab{}.
\newblock \showarticletitle{Weak SINDy for partial differential equations}.
\newblock \bibinfo{journal}{\emph{J. Comput. Phys.}}  \bibinfo{volume}{443}
  (\bibinfo{year}{2021}), \bibinfo{pages}{110525}.
\newblock


\bibitem[Meyerson et~al\mbox{.}(2024)]%
        {meyerson2024language}
\bibfield{author}{\bibinfo{person}{Elliot Meyerson}, \bibinfo{person}{Mark~J
  Nelson}, \bibinfo{person}{Herbie Bradley}, \bibinfo{person}{Adam Gaier},
  \bibinfo{person}{Arash Moradi}, \bibinfo{person}{Amy~K Hoover}, {and}
  \bibinfo{person}{Joel Lehman}.} \bibinfo{year}{2024}\natexlab{}.
\newblock \showarticletitle{Language model crossover: Variation through
  few-shot prompting}.
\newblock \bibinfo{journal}{\emph{ACM Transactions on Evolutionary Learning}}
  \bibinfo{volume}{4}, \bibinfo{number}{4} (\bibinfo{year}{2024}),
  \bibinfo{pages}{1--40}.
\newblock


\bibitem[Mirchandani et~al\mbox{.}(2023)]%
        {mirchandani2023large}
\bibfield{author}{\bibinfo{person}{Suvir Mirchandani}, \bibinfo{person}{Fei
  Xia}, \bibinfo{person}{Pete Florence}, \bibinfo{person}{Brian Ichter},
  \bibinfo{person}{Danny Driess}, \bibinfo{person}{Montserrat~Gonzalez Arenas},
  \bibinfo{person}{Kanishka Rao}, \bibinfo{person}{Dorsa Sadigh}, {and}
  \bibinfo{person}{Andy Zeng}.} \bibinfo{year}{2023}\natexlab{}.
\newblock \showarticletitle{Large Language Models as General Pattern Machines}.
  In \bibinfo{booktitle}{\emph{Conference on Robot Learning}}. PMLR,
  \bibinfo{pages}{2498--2518}.
\newblock


\bibitem[Olmo et~al\mbox{.}(2025)]%
        {olmo2025olmo}
\bibfield{author}{\bibinfo{person}{Team Olmo}, \bibinfo{person}{Allyson
  Ettinger}, \bibinfo{person}{Amanda Bertsch}, \bibinfo{person}{Bailey Kuehl},
  \bibinfo{person}{David Graham}, \bibinfo{person}{David Heineman},
  \bibinfo{person}{Dirk Groeneveld}, \bibinfo{person}{Faeze Brahman},
  \bibinfo{person}{Finbarr Timbers}, \bibinfo{person}{Hamish Ivison},
  {et~al\mbox{.}}} \bibinfo{year}{2025}\natexlab{}.
\newblock \showarticletitle{Olmo 3}.
\newblock \bibinfo{journal}{\emph{arXiv preprint arXiv:2512.13961}}
  (\bibinfo{year}{2025}).
\newblock


\bibitem[Romera-Paredes et~al\mbox{.}(2024)]%
        {romera2024mathematical}
\bibfield{author}{\bibinfo{person}{Bernardino Romera-Paredes},
  \bibinfo{person}{Mohammadamin Barekatain}, \bibinfo{person}{Alexander
  Novikov}, \bibinfo{person}{Matej Balog}, \bibinfo{person}{M~Pawan Kumar},
  \bibinfo{person}{Emilien Dupont}, \bibinfo{person}{Francisco~JR Ruiz},
  \bibinfo{person}{Jordan~S Ellenberg}, \bibinfo{person}{Pengming Wang},
  \bibinfo{person}{Omar Fawzi}, {et~al\mbox{.}}}
  \bibinfo{year}{2024}\natexlab{}.
\newblock \showarticletitle{Mathematical discoveries from program search with
  large language models}.
\newblock \bibinfo{journal}{\emph{Nature}} \bibinfo{volume}{625},
  \bibinfo{number}{7995} (\bibinfo{year}{2024}), \bibinfo{pages}{468--475}.
\newblock


\bibitem[Shojaee et~al\mbox{.}(2023)]%
        {shojaee2023transformer}
\bibfield{author}{\bibinfo{person}{Parshin Shojaee}, \bibinfo{person}{Kazem
  Meidani}, \bibinfo{person}{Amir Barati~Farimani}, {and}
  \bibinfo{person}{Chandan Reddy}.} \bibinfo{year}{2023}\natexlab{}.
\newblock \showarticletitle{Transformer-based planning for symbolic
  regression}.
\newblock \bibinfo{journal}{\emph{Advances in Neural Information Processing
  Systems}}  \bibinfo{volume}{36} (\bibinfo{year}{2023}),
  \bibinfo{pages}{45907--45919}.
\newblock


\bibitem[Shojaee et~al\mbox{.}(2025a)]%
        {shojaee2025llm}
\bibfield{author}{\bibinfo{person}{Parshin Shojaee}, \bibinfo{person}{Kazem
  Meidani}, \bibinfo{person}{Shashank Gupta}, \bibinfo{person}{Amir~Barati
  Farimani}, {and} \bibinfo{person}{Chandan~K. Reddy}.}
  \bibinfo{year}{2025}\natexlab{a}.
\newblock \showarticletitle{{LLM}-{SR}: Scientific Equation Discovery via
  Programming with Large Language Models}. In \bibinfo{booktitle}{\emph{The
  Thirteenth International Conference on Learning Representations}}.
\newblock
\urldef\tempurl%
\url{https://openreview.net/forum?id=m2nmp8P5in}
\showURL{%
\tempurl}


\bibitem[Shojaee et~al\mbox{.}(2025b)]%
        {shojaee2025llmsrbench}
\bibfield{author}{\bibinfo{person}{Parshin Shojaee}, \bibinfo{person}{Ngoc-Hieu
  Nguyen}, \bibinfo{person}{Kazem Meidani}, \bibinfo{person}{Amir~Barati
  Farimani}, \bibinfo{person}{Khoa~D Doan}, {and} \bibinfo{person}{Chandan~K
  Reddy}.} \bibinfo{year}{2025}\natexlab{b}.
\newblock \showarticletitle{Llm-srbench: A new benchmark for scientific
  equation discovery with large language models}.
\newblock \bibinfo{journal}{\emph{arXiv preprint arXiv:2504.10415}}
  (\bibinfo{year}{2025}).
\newblock


\bibitem[Team(2025)]%
        {qwen3technicalreport}
\bibfield{author}{\bibinfo{person}{Qwen Team}.}
  \bibinfo{year}{2025}\natexlab{}.
\newblock \bibinfo{title}{Qwen3 Technical Report}.
\newblock
\showeprint[arxiv]{2505.09388}~[cs.CL]
\urldef\tempurl%
\url{https://arxiv.org/abs/2505.09388}
\showURL{%
\tempurl}


\bibitem[Turc et~al\mbox{.}(2019)]%
        {turc2019}
\bibfield{author}{\bibinfo{person}{Iulia Turc}, \bibinfo{person}{Ming-Wei
  Chang}, \bibinfo{person}{Kenton Lee}, {and} \bibinfo{person}{Kristina
  Toutanova}.} \bibinfo{year}{2019}\natexlab{}.
\newblock \showarticletitle{Well-Read Students Learn Better: On the Importance
  of Pre-training Compact Models}.
\newblock \bibinfo{journal}{\emph{arXiv preprint arXiv:1908.08962v2}}
  (\bibinfo{year}{2019}).
\newblock


\bibitem[Vaswani et~al\mbox{.}(2017)]%
        {vaswani2017attention}
\bibfield{author}{\bibinfo{person}{Ashish Vaswani}, \bibinfo{person}{Noam
  Shazeer}, \bibinfo{person}{Niki Parmar}, \bibinfo{person}{Jakob Uszkoreit},
  \bibinfo{person}{Llion Jones}, \bibinfo{person}{Aidan~N Gomez},
  \bibinfo{person}{{\L}ukasz Kaiser}, {and} \bibinfo{person}{Illia
  Polosukhin}.} \bibinfo{year}{2017}\natexlab{}.
\newblock \showarticletitle{Attention is all you need}.
\newblock \bibinfo{journal}{\emph{Advances in neural information processing
  systems}}  \bibinfo{volume}{30} (\bibinfo{year}{2017}).
\newblock


\bibitem[Virtanen et~al\mbox{.}(2020)]%
        {2020SciPy-NMeth}
\bibfield{author}{\bibinfo{person}{Pauli Virtanen}, \bibinfo{person}{Ralf
  Gommers}, \bibinfo{person}{Travis~E. Oliphant}, \bibinfo{person}{Matt
  Haberland}, \bibinfo{person}{Tyler Reddy}, \bibinfo{person}{David
  Cournapeau}, \bibinfo{person}{Evgeni Burovski}, \bibinfo{person}{Pearu
  Peterson}, \bibinfo{person}{Warren Weckesser}, \bibinfo{person}{Jonathan
  Bright}, \bibinfo{person}{St{\'e}fan~J. {van der Walt}},
  \bibinfo{person}{Matthew Brett}, \bibinfo{person}{Joshua Wilson},
  \bibinfo{person}{K.~Jarrod Millman}, \bibinfo{person}{Nikolay Mayorov},
  \bibinfo{person}{Andrew R.~J. Nelson}, \bibinfo{person}{Eric Jones},
  \bibinfo{person}{Robert Kern}, \bibinfo{person}{Eric Larson},
  \bibinfo{person}{C~J Carey}, \bibinfo{person}{{\.I}lhan Polat},
  \bibinfo{person}{Yu Feng}, \bibinfo{person}{Eric~W. Moore},
  \bibinfo{person}{Jake {VanderPlas}}, \bibinfo{person}{Denis Laxalde},
  \bibinfo{person}{Josef Perktold}, \bibinfo{person}{Robert Cimrman},
  \bibinfo{person}{Ian Henriksen}, \bibinfo{person}{E.~A. Quintero},
  \bibinfo{person}{Charles~R. Harris}, \bibinfo{person}{Anne~M. Archibald},
  \bibinfo{person}{Ant{\^o}nio~H. Ribeiro}, \bibinfo{person}{Fabian Pedregosa},
  \bibinfo{person}{Paul {van Mulbregt}}, {and} \bibinfo{person}{{SciPy 1.0
  Contributors}}.} \bibinfo{year}{2020}\natexlab{}.
\newblock \showarticletitle{{{SciPy} 1.0: Fundamental Algorithms for Scientific
  Computing in Python}}.
\newblock \bibinfo{journal}{\emph{Nature Methods}}  \bibinfo{volume}{17}
  (\bibinfo{year}{2020}), \bibinfo{pages}{261--272}.
\newblock
\href{https://doi.org/10.1038/s41592-019-0686-2}{doi:\nolinkurl{10.1038/s41592-019-0686-2}}


\bibitem[Wang et~al\mbox{.}(2023)]%
        {wang2023scientific}
\bibfield{author}{\bibinfo{person}{Hanchen Wang}, \bibinfo{person}{Tianfan Fu},
  \bibinfo{person}{Yuanqi Du}, \bibinfo{person}{Wenhao Gao},
  \bibinfo{person}{Kexin Huang}, \bibinfo{person}{Ziming Liu},
  \bibinfo{person}{Payal Chandak}, \bibinfo{person}{Shengchao Liu},
  \bibinfo{person}{Peter Van~Katwyk}, \bibinfo{person}{Andreea Deac},
  {et~al\mbox{.}}} \bibinfo{year}{2023}\natexlab{}.
\newblock \showarticletitle{Scientific discovery in the age of artificial
  intelligence}.
\newblock \bibinfo{journal}{\emph{Nature}} \bibinfo{volume}{620},
  \bibinfo{number}{7972} (\bibinfo{year}{2023}), \bibinfo{pages}{47--60}.
\newblock


\bibitem[Wilstrup and Kasak(2021)]%
        {wilstrup2021symbolic}
\bibfield{author}{\bibinfo{person}{Casper Wilstrup} {and} \bibinfo{person}{Jaan
  Kasak}.} \bibinfo{year}{2021}\natexlab{}.
\newblock \showarticletitle{Symbolic regression outperforms other models for
  small data sets}.
\newblock \bibinfo{journal}{\emph{arXiv preprint arXiv:2103.15147}}
  (\bibinfo{year}{2021}).
\newblock


\bibitem[Xia et~al\mbox{.}(2025)]%
        {xia2025sr}
\bibfield{author}{\bibinfo{person}{Shijie Xia}, \bibinfo{person}{Yuhan Sun},
  {and} \bibinfo{person}{Pengfei Liu}.} \bibinfo{year}{2025}\natexlab{}.
\newblock \showarticletitle{Sr-scientist: Scientific equation discovery with
  agentic ai}.
\newblock \bibinfo{journal}{\emph{arXiv preprint arXiv:2510.11661}}
  (\bibinfo{year}{2025}).
\newblock


\bibitem[Yang et~al\mbox{.}(2023)]%
        {yang2023large}
\bibfield{author}{\bibinfo{person}{Chengrun Yang}, \bibinfo{person}{Xuezhi
  Wang}, \bibinfo{person}{Yifeng Lu}, \bibinfo{person}{Hanxiao Liu},
  \bibinfo{person}{Quoc~V Le}, \bibinfo{person}{Denny Zhou}, {and}
  \bibinfo{person}{Xinyun Chen}.} \bibinfo{year}{2023}\natexlab{}.
\newblock \showarticletitle{Large language models as optimizers}. In
  \bibinfo{booktitle}{\emph{The Twelfth International Conference on Learning
  Representations}}.
\newblock


\end{thebibliography}

\clearpage
\appendix

\section{Hyperparameters}
The hyperparamter search space for the baseline method SINDy is provided in Table \ref{tab:sindy-hyper}.

\begin{table}[h]
\small
\centering
\begin{tabular}{l l}
\toprule
\textbf{Hyperparameter} & \textbf{Possible Values} \\
\midrule
Optimizer threshold & 0.05, 0.1, 0.15 \\ \hline
Optimizer alpha     & 0.025, 0.05, 0.075 \\ \hline
Max iterations      & 20, 100 \\ \hline
Polynomial degree   & $1, 2, \dots , 10$ \\ \hline
Basis functions     & 
\begin{tabular}[t]{@{}l@{}}
\{ polynomials \} \\
\{ polynomials, sin, cos, exp\}  \\
\makecell{\{ polynomials, sin, cos, log, \\ exp, sqrt, inverse \}}
\end{tabular} \\
\bottomrule
\end{tabular}
\caption{Hyperparameter search space of SINDy.}
\label{tab:sindy-hyper}
\end{table}

\section{ODE Systems}
\label{sec:app-data}

\begin{figure*}[h]
    \centering
    \includegraphics[width=\linewidth]{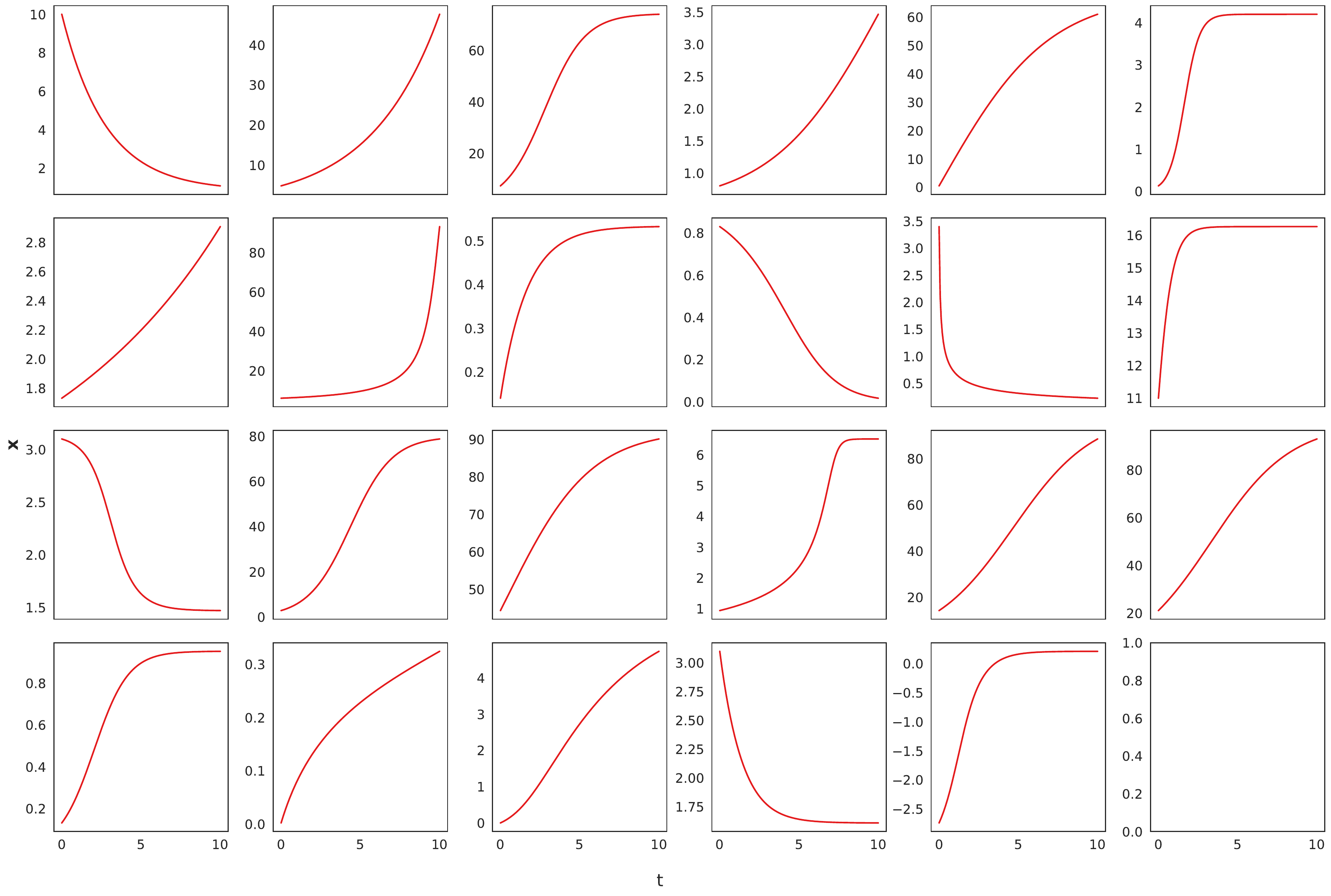}
    \caption{The training trajectories of dynamical systems with $D=1$ state variables.}
    \label{fig:traj-d1}
\end{figure*}

\begin{figure*}[h]
    \centering
    \includegraphics[width=\linewidth]{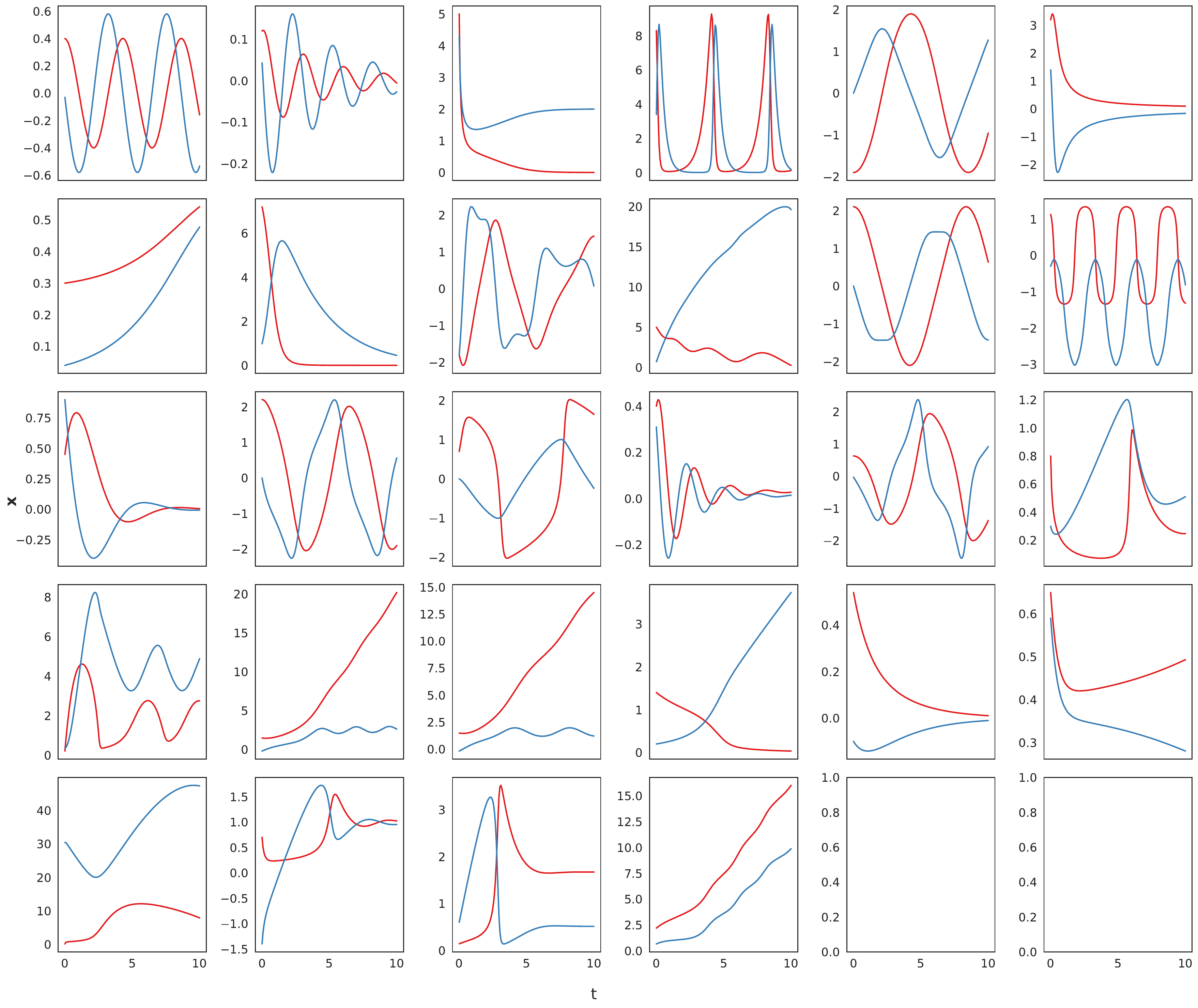}
    \caption{The training trajectories of dynamical systems with $D=2$ state variables.}
    \label{fig:traj-d2}
\end{figure*}

\begin{figure*}[h]
    \centering
    \includegraphics[width=\linewidth]{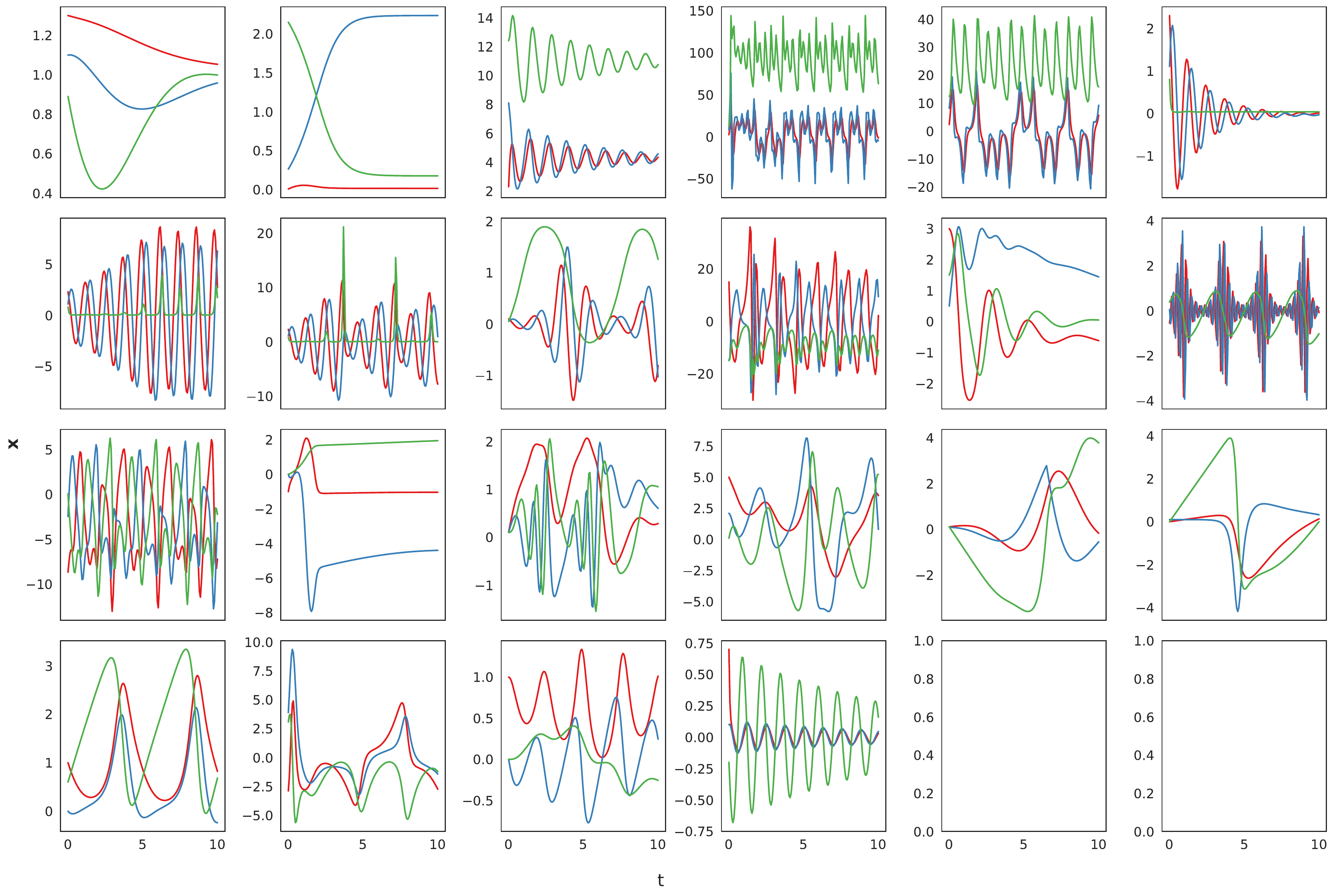}
    \caption{The training trajectories of dynamical systems with $D=3$ state variables.}
    \label{fig:traj-d3}
\end{figure*}

\begin{figure*}[h]
    \centering
    \includegraphics[width=\linewidth]{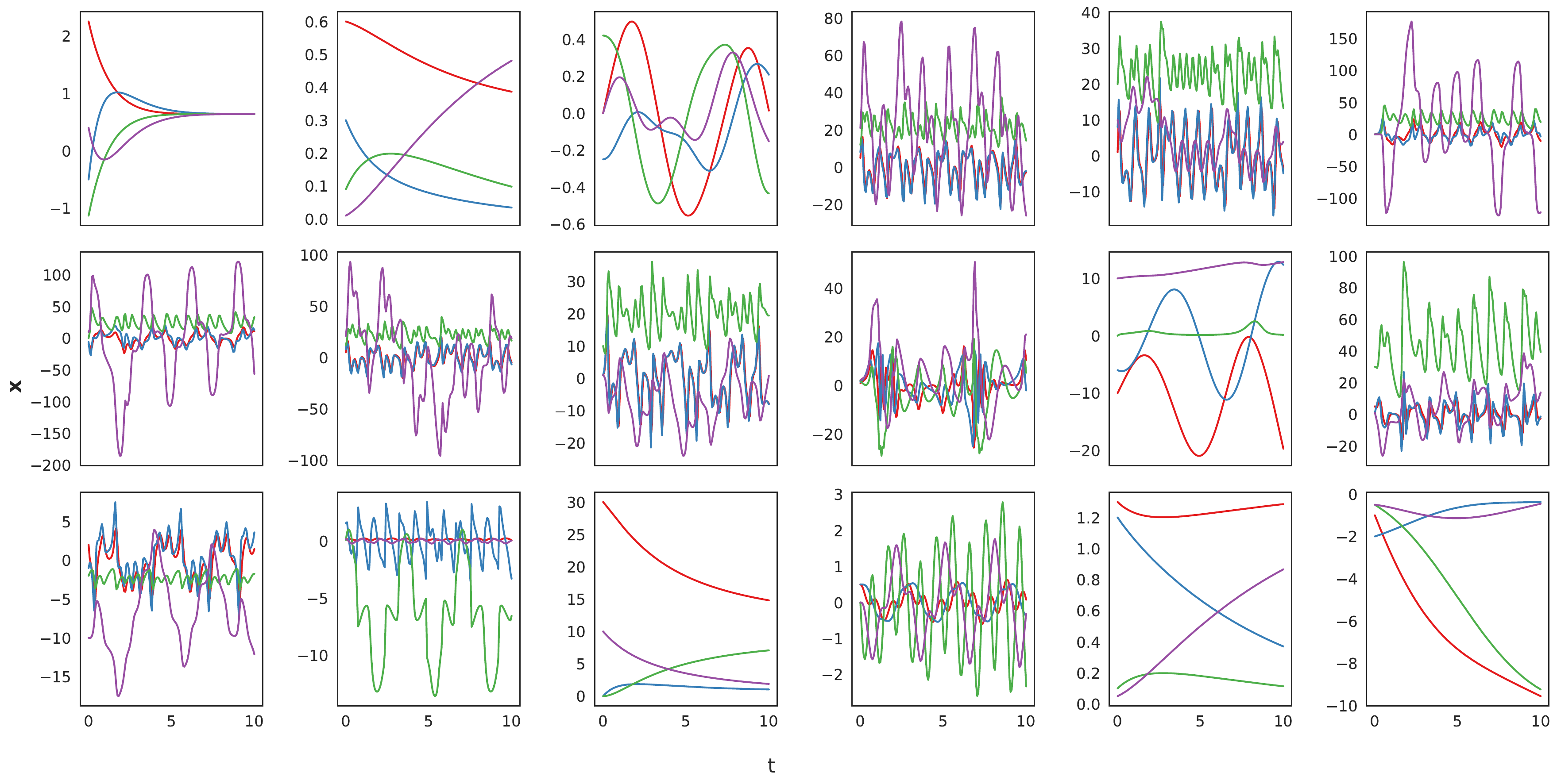}
    \caption{The training trajectories of dynamical systems with $D=4$ state variables.}
    \label{fig:traj-d4}
\end{figure*}

Tables \ref{tab:system-d1}, \ref{tab:system-d2}, \ref{tab:system-d3}, and \ref{tab:system-d4} represent the ground-truth governing equations of the dynamical systems studied in this paper. Figures \ref{fig:traj-d1}, \ref{fig:traj-d2}, \ref{fig:traj-d3}, and \ref{fig:traj-d4} visualize the trajectories of the systems, simulated from the training values. For more information about the systems including the references and system descriptions, please refer to the accompanied code supplementary material.

\begin{table*}[h]
\centering
\caption{The governing equations and the initial values of one-dimensional ($D=1$) dynamical systems.}
\label{tab:system-d1}
\renewcommand{\arraystretch}{1.2}
\begin{tabular}{|l|c|c|c|}
\hline
\textbf{Name} & \textbf{Train IV} & \textbf{Test IV} & \textbf{Equation} \\
\hline
RC-circuit
& $x_0 = 10.0$
& $x_0 = 3.54$
& $\dot{x}_0 = 0.303 - 0.361\,x_0$ \\
\hline
Population growth (naive)
& $x_0 = 4.78$
& $x_0 = 0.87$
& $\dot{x}_0 = 0.23\,x_0$ \\
\hline
Population growth (carrying capacity)
& $x_0 = 7.3$
& $x_0 = 21.0$
& $\dot{x}_0 = 0.79\,x_0\left(1 - 0.0135\,x_0\right)$ \\
\hline
RC-circuit (nonlinear resistor)
& $x_0 = 0.8$
& $x_0 = 0.02$
& $\dot{x}_0 = -0.5 + \dfrac{1}{1 + 1.65\,e^{-1.04\,x_0}}$ \\
\hline
Velocity of a falling object
& $x_0 = 0.5$
& $x_0 = 73.0$
& $\dot{x}_0 = 9.81 - 0.00212\,x_0^2$ \\
\hline
Autocatalysis
& $x_0 = 0.13$
& $x_0 = 2.24$
& $\dot{x}_0 = -0.5\,x_0^2 + 2.1\,x_0$ \\
\hline
Gompertz law (tumor growth)
& $x_0 = 1.73$
& $x_0 = 9.5$
& $\dot{x}_0 = 0.032\,x_0 \log(2.29\,x_0)$ \\
\hline
Logistic equation (Allee effect)
& $x_0 = 6.12$
& $x_0 = 2.1$
& $\dot{x}_0 = 0.14\,x_0(1 - 0.00769\,x_0)(0.227\,x_0 - 1)$ \\
\hline
Language death model
& $x_0 = 0.14$
& $x_0 = 0.55$
& $\dot{x}_0 = 0.32 - 0.6\,x_0$ \\
\hline
Refined language death model
& $x_0 = 0.83$
& $x_0 = 0.34$
& $\dot{x}_0 = -0.8\,x_0(1 - x_0)^{1.2} + 0.2\,x_0^{1.2}(1 - x_0)$ \\
\hline
Naive critical slowing down
& $x_0 = 3.4$
& $x_0 = 1.6$
& $\dot{x}_0 = -x_0^3$ \\
\hline
Photons in a laser
& $x_0 = 11.0$
& $x_0 = 1.3$
& $\dot{x}_0 = -0.111\,x_0^2 + 1.8\,x_0$ \\
\hline
Overdamped bead
& $x_0 = 3.1$
& $x_0 = 2.4$
& $\dot{x}_0 = 0.0981\,(9.7\cos x_0 - 1)\sin x_0$ \\
\hline
Budworm outbreak model
& $x_0 = 2.76$
& $x_0 = 23.3$
& $\dot{x}_0 = -\dfrac{0.9\,x_0^2}{x_0^2 + 449.44} + 0.78\,x_0(1 - 0.0123\,x_0)$ \\
\hline
Budworm outbreak (predation)
& $x_0 = 44.3$
& $x_0 = 4.5$
& $\dot{x}_0 = -\dfrac{x_0^2}{x_0^2 + 1} + 0.4\,x_0(1 - 0.0105\,x_0)$ \\
\hline
Landau equation
& $x_0 = 0.94$
& $x_0 = 1.65$
& $\dot{x}_0 = -0.001\,x_0^5 + 0.04\,x_0^3 + 0.1\,x_0$ \\
\hline
Logistic equation (harvesting)
& $x_0 = 14.3$
& $x_0 = 34.2$
& $\dot{x}_0 = 0.4\,x_0(1 - 0.01\,x_0) - 0.3$ \\
\hline
Improved logistic equation (harvesting)
& $x_0 = 21.1$
& $x_0 = 44.1$
& $\dot{x}_0 = 0.4\,x_0(1 - 0.01\,x_0) - \dfrac{0.24\,x_0}{x_0 + 50}$ \\
\hline
Improved logistic equation (dimensionless)
& $x_0 = 0.13$
& $x_0 = 0.03$
& $\dot{x}_0 = x_0(1 - x_0) - \dfrac{0.08\,x_0}{x_0 + 0.8}$ \\
\hline
Autocatalytic gene switching
& $x_0 = 0.002$
& $x_0 = 0.25$
& $\dot{x}_0 = \dfrac{x_0^2}{x_0^2 + 1} - 0.55\,x_0 + 0.1$ \\
\hline
Dimensionally reduced SIR
& $x_0 = 0.0$
& $x_0 = 0.8$
& $\dot{x}_0 = -0.2\,x_0 + 1.2 - e^{-x_0}$ \\
\hline
Protein expression
& $x_0 = 3.1$
& $x_0 = 6.3$
& $\dot{x}_0 = \dfrac{0.4\,x_0^5}{x_0^5 + 123} - 0.89\,x_0 + 1.4$ \\
\hline
Overdamped pendulum
& $x_0 = -2.74$
& $x_0 = 1.65$
& $\dot{x}_0 = 0.21 - \sin x_0$ \\
\hline
\end{tabular}
\end{table*}

\begin{table*}[h]
\centering
\caption{The governing equations and the initial values of two-dimensional ($D=2$) dynamical systems.}
\label{tab:system-d2}
\renewcommand{\arraystretch}{1.25}
\begin{adjustbox}{width=0.7\textwidth}
\begin{tabular}{|l|c|c|c|}
\hline
\textbf{Name} & \textbf{Train IV} & \textbf{Test IV} & \textbf{Equations} \\
\hline
Harmonic oscillator
& \makecell{$x_0 = 0.40$ \\ $x_1 = -0.03$}
& \makecell{$x_0 = 0.0$ \\ $x_1 = 0.20$}
& \makecell{$\dot{x}_0 = x_1$ \\ $\dot{x}_1 = -2.1\,x_0$} \\
\hline
Harmonic oscillator (damping)
& \makecell{$x_0 = 0.12$ \\ $x_1 = 0.043$}
& \makecell{$x_0 = 0.0$ \\ $x_1 = -0.30$}
& \makecell{$\dot{x}_0 = x_1$ \\ $\dot{x}_1 = -4.5\,x_0 - 0.43\,x_1$} \\
\hline
Lotka--Volterra competition
& \makecell{$x_0 = 5.0$ \\ $x_1 = 4.3$}
& \makecell{$x_0 = 2.3$ \\ $x_1 = 3.6$}
& \makecell{$\dot{x}_0 = x_0(-x_0 - 2x_1 + 3)$ \\ $\dot{x}_1 = x_1(-x_0 - x_1 + 2)$} \\
\hline
Lotka--Volterra simple
& \makecell{$x_0 = 8.3$ \\ $x_1 = 3.4$}
& \makecell{$x_0 = 0.40$ \\ $x_1 = 0.65$}
& \makecell{$\dot{x}_0 = x_0(1.84 - 1.45x_1)$ \\ $\dot{x}_1 = -x_1(3.0 - 1.62x_0)$} \\
\hline
Pendulum without friction
& \makecell{$x_0 = -1.9$ \\ $x_1 = 0.0$}
& \makecell{$x_0 = 0.30$ \\ $x_1 = 0.80$}
& \makecell{$\dot{x}_0 = x_1$ \\ $\dot{x}_1 = -0.9\sin(x_0)$} \\
\hline
Dipole fixed point
& \makecell{$x_0 = 3.2$ \\ $x_1 = 1.4$}
& \makecell{$x_0 = 1.3$ \\ $x_1 = 0.20$}
& \makecell{$\dot{x}_0 = 0.65\,x_0x_1$ \\ $\dot{x}_1 = -x_0^2 + x_1^2$} \\
\hline
Catalyzing RNA molecules
& \makecell{$x_0 = 0.30$ \\ $x_1 = 0.04$}
& \makecell{$x_0 = 0.10$ \\ $x_1 = 0.21$}
& \makecell{$\dot{x}_0 = x_0(-1.61x_0x_1 + x_1)$ \\ $\dot{x}_1 = x_1(-1.61x_0x_1 + x_0)$} \\
\hline
SIR infection
& \makecell{$x_0 = 7.2$ \\ $x_1 = 0.98$}
& \makecell{$x_0 = 20.0$ \\ $x_1 = 12.4$}
& \makecell{$\dot{x}_0 = -0.4x_0x_1$ \\ $\dot{x}_1 = 0.4x_0x_1 - 0.314x_1$} \\
\hline
Damped double well oscillator
& \makecell{$x_0 = -1.8$ \\ $x_1 = -1.8$}
& \makecell{$x_0 = 5.8$ \\ $x_1 = 0.0$}
& \makecell{$\dot{x}_0 = x_1$ \\ $\dot{x}_1 = -x_0^3 + x_0 - 0.18x_1$} \\
\hline
Glider
& \makecell{$x_0 = 5.0$ \\ $x_1 = 0.7$}
& \makecell{$x_0 = 9.81$ \\ $x_1 = -0.8$}
& \makecell{$\dot{x}_0 = -0.08x_0^2 - \sin(x_1)$ \\ $\dot{x}_1 = x_0 - \cos(x_1)/x_0$} \\
\hline
Frictionless bead
& \makecell{$x_0 = 2.1$ \\ $x_1 = 0.0$}
& \makecell{$x_0 = -1.2$ \\ $x_1 = -0.2$}
& \makecell{$\dot{x}_0 = x_1$ \\ $\dot{x}_1 = (\cos x_0 - 0.93)\sin x_0$} \\
\hline
Rotational dynamics
& \makecell{$x_0 = 1.13$ \\ $x_1 = -0.3$}
& \makecell{$x_0 = 2.4$ \\ $x_1 = 1.7$}
& \makecell{$\dot{x}_0 = \cos(x_0)\cot(x_1)$ \\ $\dot{x}_1 = (4.2\sin^2 x_1 + \cos^2 x_1)\sin x_0$} \\
\hline
Pendulum (nonlinear damping)
& \makecell{$x_0 = 0.45$ \\ $x_1 = 0.9$}
& \makecell{$x_0 = 1.34$ \\ $x_1 = -0.8$}
& \makecell{$\dot{x}_0 = x_1$ \\ $\dot{x}_1 = -0.07x_1\cos x_0 - x_1 - \sin x_0$} \\
\hline
Van der Pol oscillator
& \makecell{$x_0 = 2.2$ \\ $x_1 = 0.0$}
& \makecell{$x_0 = 0.10$ \\ $x_1 = 3.2$}
& \makecell{$\dot{x}_0 = x_1$ \\ $\dot{x}_1 = -x_0 - 0.43x_1(x_0^2 - 1)$} \\
\hline
Van der Pol (simplified)
& \makecell{$x_0 = 0.7$ \\ $x_1 = 0.0$}
& \makecell{$x_0 = -1.1$ \\ $x_1 = -0.7$}
& \makecell{$\dot{x}_0 = -1.12x_0^3 + 3.37x_0 + 3.37x_1$ \\ $\dot{x}_1 = -0.297x_0$} \\
\hline
Glycolytic oscillator
& \makecell{$x_0 = 0.40$ \\ $x_1 = 0.31$}
& \makecell{$x_0 = 0.20$ \\ $x_1 = -0.7$}
& \makecell{$\dot{x}_0 = x_0^2x_1 - x_0 + 2.4x_1$ \\ $\dot{x}_1 = -x_0^2x_1 - 2.4x_0 + 0.07$} \\
\hline
Duffing equation
& \makecell{$x_0 = 0.63$ \\ $x_1 = -0.03$}
& \makecell{$x_0 = 0.20$ \\ $x_1 = 0.20$}
& \makecell{$\dot{x}_0 = x_1$ \\ $\dot{x}_1 = -x_0 + 0.886x_1(1 - x_0^2)$} \\
\hline
Cell cycle model
& \makecell{$x_0 = 0.8$ \\ $x_1 = 0.3$}
& \makecell{$x_0 = 0.02$ \\ $x_1 = 1.2$}
& \makecell{$\dot{x}_0 = -x_0 + 15.3(-x_0 + x_1)(x_0^2 + 0.001)$ \\ $\dot{x}_1 = 0.3 - x_0$} \\
\hline
CDIMA reaction
& \makecell{$x_0 = 0.20$ \\ $x_1 = 0.35$}
& \makecell{$x_0 = 3.0$ \\ $x_1 = 7.8$}
& \makecell{$\dot{x}_0 = -\frac{4x_0x_1}{x_0^2+1} - x_0 + 8.9$ \\ $\dot{x}_1 = 1.4x_0\!\left(1 - \frac{x_1}{x_0^2+1}\right)$} \\
\hline
Driven pendulum (linear damping)
& \makecell{$x_0 = 1.47$ \\ $x_1 = -0.2$}
& \makecell{$x_0 = -1.9$ \\ $x_1 = 0.03$}
& \makecell{$\dot{x}_0 = x_1$ \\ $\dot{x}_1 = -0.64x_1 - \sin x_0 + 1.67$} \\
\hline
Driven pendulum (quadratic damping)
& \makecell{$x_0 = 1.47$ \\ $x_1 = -0.2$}
& \makecell{$x_0 = -1.9$ \\ $x_1 = 0.03$}
& \makecell{$\dot{x}_0 = x_1$ \\ $\dot{x}_1 = -0.64x_1|x_1| - \sin x_0 + 1.67$} \\
\hline
Gray--Scott model
& \makecell{$x_0 = 1.4$ \\ $x_1 = 0.2$}
& \makecell{$x_0 = 0.32$ \\ $x_1 = 0.64$}
& \makecell{$\dot{x}_0 = -x_0x_1^2 - 0.5x_0 + 0.5$ \\ $\dot{x}_1 = x_0x_1^2 - 0.02x_1$} \\
\hline
Interacting bar magnets
& \makecell{$x_0 = 0.54$ \\ $x_1 = -0.1$}
& \makecell{$x_0 = 0.43$ \\ $x_1 = 1.21$}
& \makecell{$\dot{x}_0 = -\sin x_0 + 0.33\sin(x_0-x_1)$ \\ $\dot{x}_1 = -\sin x_1 - 0.33\sin(x_0-x_1)$} \\
\hline
Binocular rivalry model
& \makecell{$x_0 = 0.65$ \\ $x_1 = 0.59$}
& \makecell{$x_0 = 3.2$ \\ $x_1 = 10.3$}
& \makecell{$\dot{x}_0 = -x_0 + \frac{1}{0.247e^{4.89x_1}+1}$ \\ $\dot{x}_1 = -x_1 + \frac{1}{0.247e^{4.89x_0}+1}$} \\
\hline
Bacterial respiration model
& \makecell{$x_0 = 0.1$ \\ $x_1 = 30.4$}
& \makecell{$x_0 = 13.2$ \\ $x_1 = 5.21$}
& \makecell{$\dot{x}_0 = -\frac{x_0x_1}{0.48x_0^2+1} - x_0 + 18.3$ \\ $\dot{x}_1 = -\frac{x_0x_1}{0.48x_0^2+1} + 11.23$} \\
\hline
Brusselator
& \makecell{$x_0 = 0.7$ \\ $x_1 = -1.4$}
& \makecell{$x_0 = 2.1$ \\ $x_1 = 1.3$}
& \makecell{$\dot{x}_0 = 3.1x_0^2x_1 - 4.03x_0 + 1$ \\ $\dot{x}_1 = -3.1x_0^2x_1 + 3.03x_0$} \\
\hline
Schnackenberg model
& \makecell{$x_0 = 0.14$ \\ $x_1 = 0.6$}
& \makecell{$x_0 = 1.5$ \\ $x_1 = 0.9$}
& \makecell{$\dot{x}_0 = x_0^2x_1 - x_0 + 0.24$ \\ $\dot{x}_1 = -x_0^2x_1 + 1.43$} \\
\hline
Oscillator death model
& \makecell{$x_0 = 2.2$ \\ $x_1 = 0.67$}
& \makecell{$x_0 = 0.03$ \\ $x_1 = -0.12$}
& \makecell{$\dot{x}_0 = \sin x_1\cos x_0 + 1.432$ \\ $\dot{x}_1 = \sin x_1\cos x_0 + 0.972$} \\
\hline
\end{tabular}
\end{adjustbox}
\end{table*}

\begin{table*}[h]
\centering
\caption{The governing equations and the initial values of three-dimensional ($D=3$) dynamical systems.}
\label{tab:system-d3}
\renewcommand{\arraystretch}{1.3}
\begin{adjustbox}{width=0.8\textwidth}
\begin{tabular}{|l|c|c|l|}
\hline
\textbf{Name} & \textbf{Train IV} & \textbf{Test IV} & \textbf{Equations} \\
\hline
Maxwell-Bloch equations & $[1.3, 1.1, 0.89]$ & $[0.89, 1.3, 1.1]$ & \makecell{$\dot{x}_0 = -0.1x_0 + 0.1x_1$ \\ $\dot{x}_1 = 0.21x_0x_2 - 0.21x_1$ \\ $\dot{x}_2 = -1.054x_0x_1 - 0.34x_2 + 1.394$} \\
\hline
Apoptosis model & $[0.005, 0.26, 2.15]$ & $[0.248, 0.097, 0.003]$ & \makecell{$\dot{x}_0 = -0.4x_0x_1/(x_0 + 0.1) - 0.05x_0 + 0.1$ \\ $\dot{x}_1 = -7.95x_0x_1/(x_1 + 2.0) - 0.2x_1/(x_1 + 0.1) + 0.6x_2(x_1 + 0.1)$ \\ $\dot{x}_2 = 7.95x_0x_1/(x_1 + 2.0) + 0.2x_1/(x_1 + 0.1) - 0.6x_2(x_1 + 0.1)$} \\
\hline
Lorenz equations periodic & $[2.3, 8.1, 12.4]$ & $[10.0, 20.0, 30.0]$ & \makecell{$\dot{x}_0 = -5.1x_0 + 5.1x_1$ \\ $\dot{x}_1 = -x_0x_2 + 12.0x_0 - x_1$ \\ $\dot{x}_2 = x_0x_1 - 1.67x_2$} \\
\hline
\makecell[l]{Lorenz equations \\ complex periodic} & $[2.3, 8.1, 12.4]$ & $[10.0, 20.0, 30.0]$ & \makecell{$\dot{x}_0 = -10.0x_0 + 10.0x_1$ \\ $\dot{x}_1 = -x_0x_2 + 99.96x_0 - x_1$ \\ $\dot{x}_2 = x_0x_1 - 2.67x_2$} \\
\hline
Lorenz equations chaotic & $[2.3, 8.1, 12.4]$ & $[10.0, 20.0, 30.0]$ & \makecell{$\dot{x}_0 = -10.0x_0 + 10.0x_1$ \\ $\dot{x}_1 = -x_0x_2 + 28.0x_0 - x_1$ \\ $\dot{x}_2 = x_0x_1 - 2.67x_2$} \\
\hline
Rössler fixed point & $[2.3, 1.1, 0.8]$ & $[-0.1, 4.1, -2.1]$ & \makecell{$\dot{x}_0 = -5.0x_1 - 5.0x_2$ \\ $\dot{x}_1 = 5.0x_0 - 1.0x_1$ \\ $\dot{x}_2 = 5.0x_2(x_0 - 5.7) + 1.0$} \\
\hline
\makecell[l]{Rössler attractor \\ periodic} & $[2.3, 1.1, 0.8]$ & $[-0.1, 4.1, -2.1]$ & \makecell{$\dot{x}_0 = -5.0x_1 - 5.0x_2$ \\ $\dot{x}_1 = 5.0x_0 + 0.5x_1$ \\ $\dot{x}_2 = 5.0x_2(x_0 - 5.7) + 1.0$} \\
\hline
\makecell[l]{Rössler attractor \\ chaotic} & $[2.3, 1.1, 0.8]$ & $[-0.1, 4.1, -2.1]$ & \makecell{$\dot{x}_0 = -5.0x_1 - 5.0x_2$ \\ $\dot{x}_1 = 5.0x_0 + 1.0x_1$ \\ $\dot{x}_2 = 5.0x_2(x_0 - 5.7) + 1.0$} \\
\hline
Aizawa attractor & $[0.1, 0.05, 0.05]$ & $[-0.3, 0.2, 0.1]$ & \makecell{$\dot{x}_0 = x_0(x_2 - 0.7) - 3.5x_1$ \\ $\dot{x}_1 = 3.5x_0 + x_1(x_2 - 0.7)$ \\ $\dot{x}_2 = 0.1x_0^3x_2 - 0.33x_2^3 + 0.95x_2 - (x_0^2 + x_1^2)(0.25x_2 + 1) + 0.65$} \\
\hline
Chen-Lee attractor & $[15, -15, -15]$ & $[8, 14, -10]$ & \makecell{$\dot{x}_0 = 5.0x_0 - x_1x_2$ \\ $\dot{x}_1 = x_0x_2 - 10.0x_1$ \\ $\dot{x}_2 = 0.33x_0x_1 - 3.8x_2$} \\
\hline
\makecell[l]{Gissinger chaotic \\ reversals} & $[3, 0.5, 1.5]$ & $[1.0, 1.0, 0]$ & \makecell{$\dot{x}_0 = 0.119x_0 - x_1x_2$ \\ $\dot{x}_1 = x_0x_2 - 0.1x_1$ \\ $\dot{x}_2 = x_0x_1 - x_2 + 0.9$} \\
\hline
\makecell[l]{Nonlinear oscillator \\ with 3 states} & \makecell{$[-0.56, 0.05,0.38]$} & $[-0.1, 0.4, 1]$ & \makecell{$\dot{x}_0 = 3x_0x_2 + 0.4x_0 - 20.25x_1 + 1.6x_2(x_0^2 + x_1^2)$ \\ $\dot{x}_1 = 20.25x_0 + 3x_1x_2 + 0.4x_1$ \\ $\dot{x}_2 = -0.44x_0^2 - 0.44x_1^2 - 0.4x_2^3 - x_2^2 + 1.7$} \\
\hline
Sprott chaotic system & \makecell{$[-8.68, -2.47,$ \\ $0.07]$} & $[-2.1, 2, 0]$ & \makecell{$\dot{x}_0 = -1.4x_0 - x_1^2 - 4x_1 - 4x_2$ \\ $\dot{x}_1 = -4x_0 - 1.4x_1 - x_2^2 - 4x_2$ \\ $\dot{x}_2 = -x_0^2 - 4x_0 - 4x_1 - 1.4x_2$} \\
\hline
Hindmarsh-Rose model & $[-1, 0, 0]$ & $[-0.5, 0.2, -1]$ & \makecell{$\dot{x}_0 = -x_0^3 + 3x_0^2 + x_1 - x_2 + 2$ \\ $\dot{x}_1 = -5x_0^2 - x_1 + 1$ \\ $\dot{x}_2 = 0.4x_0 - 0.1x_2 + 0.64$} \\
\hline
Lorenz-84 system & $[0.1, 0.1, 0.1]$ & $[-0.1, 0.2, 0.4]$ & \makecell{$\dot{x}_0 = -0.25x_0 - x_1^2 - x_2^2 + 1.7115$ \\ $\dot{x}_1 = x_0x_1 - 4.0x_0x_2 - x_1 + 1.287$ \\ $\dot{x}_2 = 4.0x_0x_1 + x_0x_2 - x_2$} \\
\hline
Diffusionless Lorenz system & $[5, 2.1, 0.1]$ & $[-2, -1.2, 0.4]$ & \makecell{$\dot{x}_0 = -x_0 + x_1$ \\ $\dot{x}_1 = -x_0x_2$ \\ $\dot{x}_2 = x_0x_1 - 4.7$} \\
\hline
\makecell[l]{Sprott multifractal \\ system} & $[0.1, 0.1, 0.1]$ & $[-1.1, 0.2, -4]$ & \makecell{$\dot{x}_0 = x_1$ \\ $\dot{x}_1 = -x_0 - x_1 \text{ sgn}(x_2)$ \\ $\dot{x}_2 = x_1^2 - \exp(-x_0^2)$} \\
\hline
Nosé-Hoover system & $[0.0, 0.1, 0.0]$ & $[1.0, -1.0, 0.0]$ & \makecell{$\dot{x}_0 = x_1$ \\ $\dot{x}_1 = -x_0 + x_1x_2$ \\ $\dot{x}_2 = 1 - x_1^2$} \\
\hline
Rikitake's dynamo & $[1.0, 0.0, 0.6]$ & $[-1.0, 0.0, -0.6]$ & \makecell{$\dot{x}_0 = -1.0x_0 + x_1x_2$ \\ $\dot{x}_1 = x_0(x_2 - 1.0) - 1.0x_1$ \\ $\dot{x}_2 = -x_0x_1 + 1$} \\
\hline
Li system & \makecell{$[-2.9, 3.89,$ \\ $3.07]$} & $[1.3, -4.34, 9]$ & \makecell{$\dot{x}_0 = 1.0x_0 + x_1x_2 + x_1$ \\ $\dot{x}_1 = -x_0x_2 + x_1x_2$ \\ $\dot{x}_2 = -1.0x_0x_1 - x_2 + 1.0$} \\
\hline
\makecell[l]{Sprott dissipative- \\ conservative system} & $[1.0, 0.0, 0.0]$ & $[2.0, 0.0, 0.0]$ & \makecell{$\dot{x}_0 = 2.0x_0x_1 + x_0x_2 + x_1$ \\ $\dot{x}_1 = -2x_0^2 + 1.0x_1x_2 + 1$ \\ $\dot{x}_2 = -x_0^2 + 1.0x_0 - x_1^2$} \\
\hline
Chua chaotic circuit & $[0.7, 0.1, -0.2]$ & $[3.23, 0.1, -7.2]$ & \makecell{$\dot{x}_0 = -11.14x_0 + 15.6x_1 + 3.34(|x_0-1| - |x_0+1|)$ \\ $\dot{x}_1 = x_0 - x_1 + x_2, \quad \dot{x}_2 = -25.58x_1$} \\
\hline
\end{tabular}
\end{adjustbox}
\end{table*}

\begin{table*}[h]
\centering
\small
\caption{The governing equations and the initial values of four-dimensional ($D=4$) dynamical systems.}
\label{tab:system-d4}
\renewcommand{\arraystretch}{1.4}
\begin{adjustbox}{max width=\textwidth}
\begin{tabular}{|l|c|c|l|}
\hline
\textbf{Name} & \textbf{Train IV} & \textbf{Test IV} & \textbf{Equations} \\
\hline
\makecell[l]{Binocular rivalry \\ adaptation} & \makecell{$[2.25,-0.5,$\\$-1.13,0.4]$} & \makecell{$0.34, -0.43,$\\$-0.86,0.04]$} & \makecell{$\dot{x}_0 = -x_0 + (0.247e^{0.4x_1 + 0.89x_2} + 1)^{-1}, \dot{x}_1 = x_0 - x_1$ \\ $\dot{x}_2 = -x_2 + (0.247e^{0.89x_0 + 0.4x_3} + 1)^{-1}, \dot{x}_3 = x_2 - x_3$} \\
\hline
SEIR infection & $[0.6, 0.3, 0.09, 0.01]$ & $[0.4, 0.3, 0.25, 0.05]$ & \makecell{$\dot{x}_0 = -0.28x_0x_2, \quad \dot{x}_1 = 0.28x_0x_2 - 0.47x_1$ \\ $\dot{x}_2 = 0.47x_1 - 0.3x_2, \quad \dot{x}_3 = 0.3x_2$} \\
\hline
Hénon-Heiles & $[0, -0.25, 0.42, 0]$ & $[0.3, -0.2, 0.28, 0.01]$ & \makecell{$\dot{x}_0 = x_2, \quad \dot{x}_1 = x_3$ \\ $\dot{x}_2 = -2x_0x_1 - x_0, \quad \dot{x}_3 = -x_0^2 + x_1^2 - x_1$} \\
\hline
\makecell[l]{Hyperchaotic Lu system \\ (Bo-Cheng \& Zhong)} & $[5, 8, 12, 21]$ & $[9, 8, 5.8, 11]$ & \makecell{$\dot{x}_0 = -36x_0 + 36x_1 + x_3, \quad \dot{x}_1 = -x_0x_2 + 20.5x_1$ \\ $\dot{x}_2 = x_0x_1 - 3x_2, \quad \dot{x}_3 = 21x_0 - 0.1x_1x_2$} \\
\hline
\makecell[l]{Cai-Huang hyperchaotic \\ finance system} & $[1, 8, 20, 10]$ & $[2, 5, 15, 17]$ & \makecell{$\dot{x}_0 = -27.5x_0 + 27.5x_1, \quad \dot{x}_2 = x_1^2 - 2.9x_2$ \\ $\dot{x}_1 = -x_0x_2 + 3x_0 + 19.3x_1 + x_3, \quad \dot{x}_3 = -3.3x_0$} \\
\hline
\makecell[l]{Hyperchaotic Lorenz system \\ (Hussain-Gondal-Hussain)} & $[0.1, 0.1, 0.1, 0.1]$ & $[-10, -6, 0, 10]$ & \makecell{$\dot{x}_0 = -10x_0 + 10x_1 + x_3, \quad \dot{x}_1 = x_0(28 - x_2) - x_1$ \\ $\dot{x}_2 = x_0x_1 - 2.67x_2, \quad \dot{x}_3 = -x_0x_2 + 1.3x_3$} \\
\hline
\makecell[l]{Hyperchaotic Lorenz system \\ (Wang-Wang)} & $[-10, -6, 0, 10]$ & $[4, 0.2, 0, -1]$ & \makecell{$\dot{x}_0 = -10x_0 + 10x_1 + x_3, \quad \dot{x}_1 = x_0(28 - x_2) - x_1$ \\ $\dot{x}_2 = x_0x_1 - 2.67x_2, \quad \dot{x}_3 = -x_1x_2 - x_3$} \\
\hline
\makecell[l]{Hyperchaotic Lu system \\ (Chen-Lu-Lü-Yu)} & $[5, 8, 12, 21]$ & $[0.2, -0.2, 0.2, 0.1]$ & \makecell{$\dot{x}_0 = -36x_0 + 36x_1 + x_3, \quad \dot{x}_1 = -x_0x_2 + 20x_1$ \\ $\dot{x}_2 = x_0x_1 - 3x_2, \quad \dot{x}_3 = x_0x_2 + 1.3x_3$} \\
\hline
Hyperchaotic Pang system & $[1, 1, 10, 1]$ & $[0.2, -0.2, -0.2, -0.1]$ & \makecell{$\dot{x}_0 = -36x_0 + 36x_1, \quad \dot{x}_1 = -x_0x_2 + 20x_1 + x_3$ \\ $\dot{x}_2 = x_0x_1 - 3x_2, \quad \dot{x}_3 = -2x_0 - 2x_1$} \\
\hline
Hyperchaotic Qi system & $[1, 1.5, 2, 2.2]$ & $[0.1, -3.1, 0.2, -0.1]$ & \makecell{$\dot{x}_0 = -5x_0 + x_1x_2 + 5x_1, \quad \dot{x}_1 = -x_0x_2 + 2.4x_0 + 2.4x_1$ \\ $\dot{x}_2 = x_0x_1 - 1.3x_2 - 3.3x_3, \quad \dot{x}_3 = x_0x_1 + 3x_2 - 0.8x_3$} \\
\hline
Hyperchaotic Rössler system & $[-10, -6, 0, 10]$ & $[-0.2, -6, 0, 11]$ & \makecell{$\dot{x}_0 = -x_1 - x_2, \quad \dot{x}_1 = x_0 + 0.25x_1 + x_3$ \\ $\dot{x}_2 = x_0x_2 + 3, \quad \dot{x}_3 = -0.5x_2 + 0.05x_3$} \\
\hline
Hyperchaotic Wang system & $[5, 1, 30, 1]$ & $[1, 1, 1, 1]$ & \makecell{$\dot{x}_0 = -10x_0 + 10x_1, \quad \dot{x}_1 = -x_0x_2 + 40x_0 + x_3$ \\ $\dot{x}_2 = 4x_0^2 - 2.5x_2, \quad \dot{x}_3 = -10.6x_0$} \\
\hline
Hyperchaotic Xu system & $[2, -1, -2, -10]$ & $[-1, 1, 1, -1]$ & \makecell{$\dot{x}_0 = -10x_0 + 10x_1 + x_3, \quad \dot{x}_1 = 16x_0x_2 + 40x_0$ \\ $\dot{x}_2 = -x_0x_1 - 2.5x_2, \quad \dot{x}_3 = x_0x_2 - 2x_1$} \\
\hline
Swinging Atwood's machine & \makecell{$[0.11, 1.57,$\\$0.11,0.18]$} & $[0.6, 2.9, -0.1, 0.1]$ & \makecell{$\dot{x}_0 = 0.33x_1, \quad \dot{x}_1 = 9.81\cos(x_2) - 19.61 + x_3^2/x_0^3$ \\ $\dot{x}_2 = x_3/x_0^2, \quad \dot{x}_3 = -9.81x_0\sin(x_2)$} \\
\hline
\makecell[l]{Polymer DA cross-linking \\ and rDA de-cross-linking} & $[30, 0, 0, 10]$ & $[0.01, 0.08, 0.03, 0.03]$ & \makecell{$\dot{x}_0 = -0.03x_0x_1 - 0.01x_0x_3 + 0.08x_1 + 0.03x_2$ \\ $\dot{x}_1 = -0.03x_0x_1 + 0.01x_0x_3 - 0.08x_1 + 0.03x_2$ \\ $\dot{x}_2 = 0.03x_0x_1 - 0.03x_2, \quad \dot{x}_3 = -0.01x_0x_3 + 0.08x_1$} \\
\hline
Double Pendulum & $[0.5, 0.5, 0, 0]$ & $[0.1, 0.2, 0, 0]$ & \makecell{$\dot{x}_0 = x_2, \quad \dot{x}_1 = x_3, \quad \Delta = x_0 - x_1$ \\ $\dot{x}_2 = \frac{2(-0.55x_2^2\sin\Delta\cos\Delta - 2.09x_3^2\sin\Delta - 22.56\sin x_0 + 10.79\sin x_1\cos\Delta)}{2.3 - 1.1\cos^2\Delta}$ \\ $\dot{x}_3 = \frac{0.53(1.15x_2^2\sin\Delta + 2.09x_3^2\sin\Delta\cos\Delta + 22.56\sin x_0\cos\Delta - 22.56\sin x_1)}{2.3 - 1.1\cos^2\Delta}$} \\
\hline
\makecell[l]{Enzyme kinetics \\ (Michaelis-Menten)} & $[1.3, 1.2, 0.1, 0.05]$ & $[0.1, 0.2, 0.3, 0.4]$ & \makecell{$\dot{x}_0 = -0.1x_0x_1 + 0.52x_2, \quad \dot{x}_1 = -0.1x_0x_1 + 0.02x_2$ \\ $\dot{x}_2 = 0.1x_0x_1 - 0.52x_2, \quad \dot{x}_3 = 0.5x_2$} \\
\hline
\makecell[l]{Two-mass spring system \\ with damping and force} & $[-1, -2, -0.5, -0.5]$ & $[-0.1, 0.3, 0.3, -0.4]$ & \makecell{$\dot{x}_0 = x_1, \quad \dot{x}_2 = x_3$ \\ $\dot{x}_1 = -0.09x_0 - 0.13x_1 + 0.09x_2 + 0.13x_3$ \\ $\dot{x}_3 = 0.11x_0 + 0.16x_1 - 0.11x_2 - 0.16x_3 + 0.2$} \\
\hline
\end{tabular}
\end{adjustbox}
\end{table*}

\clearpage

\end{document}